\documentclass[preprints,article,accept,moreauthors,pdftex]{Definitions/mdpi}

\usepackage{multirow}
\usepackage{color}
\usepackage{todonotes}
\usepackage{comment}
\usepackage{ragged2e}
\usepackage{siunitx}
\firstpage{1} 
\pubvolume{xx}
\issuenum{1}
\articlenumber{5}
\pubyear{2019}
\copyrightyear{2019}
\history{Received: date; Accepted: date; Published: date}

\Title{deepNIR: Datasets for generating synthetic NIR images and improved fruit detection system using deep learning techniques}

\Author{Inkyu Sa $^{1,}$*\orcidA{}, JongYoon Lim $^{2}$, Ho Seok Ahn $^{2}$ and Bruce MacDonald $^{2}$}

\AuthorNames{Inkyu Sa, JongYoon Lim, Ho Seok Ahn and Bruce MacDonald} 

\address{%
$^{1}$ \quad CSIRO Data61, Robotics and Autonomous Systems Group, Robot perception team
; inkyu.sa@csiro.au \\
$^{2}$ \quad CARES, Department of Electrical, Computer and Software Engineering, University of Auckland; {jy.lim, hs.ahn, b.macdonald}@auckland.ac.nz
}

\corres{Correspondence: inkyu.sa@csiro.au; Tel.: +61 7 3327 4317 (S. I.)}

\firstnote{Current address: Affiliation 3} 

\abstract{This paper presents datasets utilised for synthetic near-infrared (NIR) image generation and bounding-box level fruit detection systems. A high-quality dataset is one of the essential building blocks that can lead to success in model generalisation and the deployment of data-driven deep neural networks. In particular, synthetic data generation tasks often require more training samples than other supervised approaches.
Therefore, in this paper, we share the NIR+RGB datasets that are re-processed from two public datasets (i.e., nirscene and SEN12MS), expanded our previous study, deepFruits, and our novel NIR+RGB sweet pepper(capsicum) dataset. We oversampled from the original nirscene dataset at 10, 100, 200, and 400 ratios that yielded a total of 127\unit{k} pairs of images. From the SEN12MS satellite multispectral dataset, we selected Summer (45\unit{k}) and All seasons (180\unit{k}) subsets and applied a simple yet important conversion; digital number (DN) to pixel value conversion followed by image standardisation. Our sweet pepper dataset consists of 1615 pairs of NIR+RGB images that were collected from commercial farms. We quantitatively and qualitatively demonstrate that these NIR+RGB datasets are sufficient to be used for synthetic NIR image generation. We achieved Frechet Inception Distance (FID) of 11.36, 26.53, and 40.15 for nirscene1, SEN12MS, and sweet pepper datasets respectively. In addition, we release manual annotations of \textbf{11} fruit bounding boxes that can be exported as various formats using cloud service. Four newly added fruits [blueberry, cherry, kiwi, and wheat] compound 11 novel bounding box datasets on top of our previous work presented in the deepFruits project [apple, avocado, capsicum, mango, orange, rockmelon, strawberry]. The total number of bounding box instances of the dataset is 162\unit{k} and it is ready to use from cloud service. For the evaluation of the dataset, Yolov5 single stage detector is exploited and reported impressive mean-average-precision, $\text{mAP}_{\mbox{\tiny{[0.5:0.95]}}}$ results of [min:0.49, max:0.812]. We hope these datasets are useful and serve as a baseline for the future studies. The datasets are available from \textbf{http://tiny.one/deepNIR}}


\def\fps@figure{htp}
\def\fps@table{htp}

\newcommand{\bi}{\begin{itemize}}
\newcommand{\ei}{\end{itemize}}

\newcommand{\bfig}{\begin{figure}}
\newcommand{\efig}{\end{figure}}

\newcommand{\benum}{\begin{enumerate}}
\newcommand{\eenum}{\end{enumerate}}

\newcommand{\be}{\begin{equation}}
\newcommand{\ee}{\end{equation}}

\newcommand{\ba}{\begin{eqnarray}}
\newcommand{\ea}{\end{eqnarray}}

\newcommand{\unit}[1]{\mbox{$\rm \,#1$}}

\begin{document}
\section{Introduction}
The recent advances in data-driven Machine Learning (ML) techniques have been unlocked and achieved impressive outcomes in industrial research sectors and even our daily life. As exemplified applications such as Autonomous driving\cite{Sun2019-jr}, Natural Language Processing (NLP)\cite{Vaswani2017-vx}, synthetic visual data generation\cite{Korshunov2018-ep}, protein structure prediction\cite{Jumper2021-dz}, and nuclear fusion reactor control\cite{Degrave2022-ru}, it is very exciting to see what else ML can learn and how much it will bring impact to our future life.

\begin{figure}
\centering
\includegraphics[width=\textwidth]{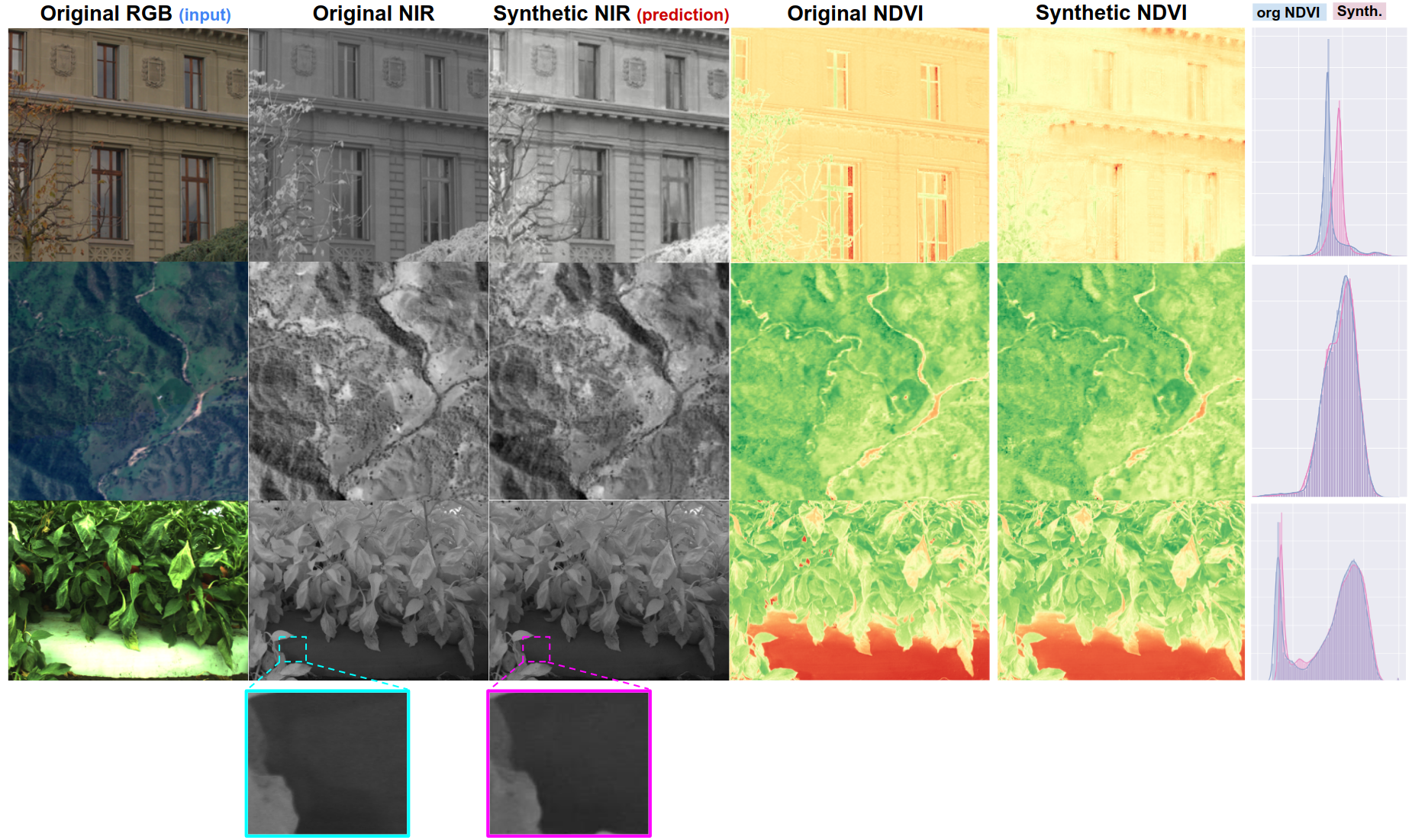}
\caption{Each row indicates a data sample and the corresponding output drawn from nirscene, SEN12MS, and capsicum datasets respectively. The 1st and 2nd columns are RGB+NIR image pair used for model training, 3rd is a NIR prediction given the RGB image and their normalised difference vegetation index(NDVI) for the rest of the columns. This figure is best viewed in colour.}
\label{fig:front}
\end{figure}

In this paper, we are interested in bringing these data-driven ML technologies to the agriculture sector to take considerable advantages in core agricultural tasks such as vegetation segmentation and fruit detection. Toward this, we adopt one of the ML techniques, synthetic image generation and data-driven object detection, to see how much we can obtain improvement. Especially, we first focus on generating synthetic near-infrared (NIR) images that can then be used for object detection as auxiliary information, as shown in Figure\ref{fig:front}.

Within the agriculture domain, NIR information ($\lambda\sim750-850$\unit{nm}) has played a pivotal role in various tasks since 1970. One of the most important contributions is enabling vegetation indices (NDVI) with a simple and fast closed-form\cite{Rouse1973-oz}, and this still now sets a stepping stone for many other advanced indices such as the Enhanced vegetation index (EVI) or Normalized difference water index (NDWI). Analogous to thermal spectrum which allows measuring beyond the visible range and brings significant salient features,  the NIR spectrum enables observing plants' chlorophyll responses (mainly from leaves). This information is crucial for agronomists to phenotype vegetation's status and conditions.


In order to synthesise NIR information from RGB input, it is necessary to properly approximate a highly nonlinear mapping, $f_{\theta}$ such that $f_{\theta}: \{{I_{\mbox{\tiny{RGB}}}}\} \mapsto I_{\mbox{\tiny{NIR}}}$ where $\theta$ is unknown parameters (e.g., neural networks' parameters). The nonlinearity stemms from incident lighting sources, surface reflectances, intrinsic and extrinsic camera inherent characteristics and many other factors. Hence, it is one of the challenges to estimate the global optimal solution that guarantees convergence. Instead, we attempt to learn the mapping in a data-driven, unsupervised manner that does not require manual annotations or labels. To do this, we set the objective function that minimises differences between synthetic and original NIR images given RGB images. This idea is straightforward, and there are already several previous studies in generating not only NIR \cite{An2019-xl, Yuan2020-ma} but also thermal\cite{Bhat2020-fg} and depth\cite{Saxena2005-bs, Zheng2018-uk}.

\begin{figure}
\centering
\includegraphics[width=\textwidth]{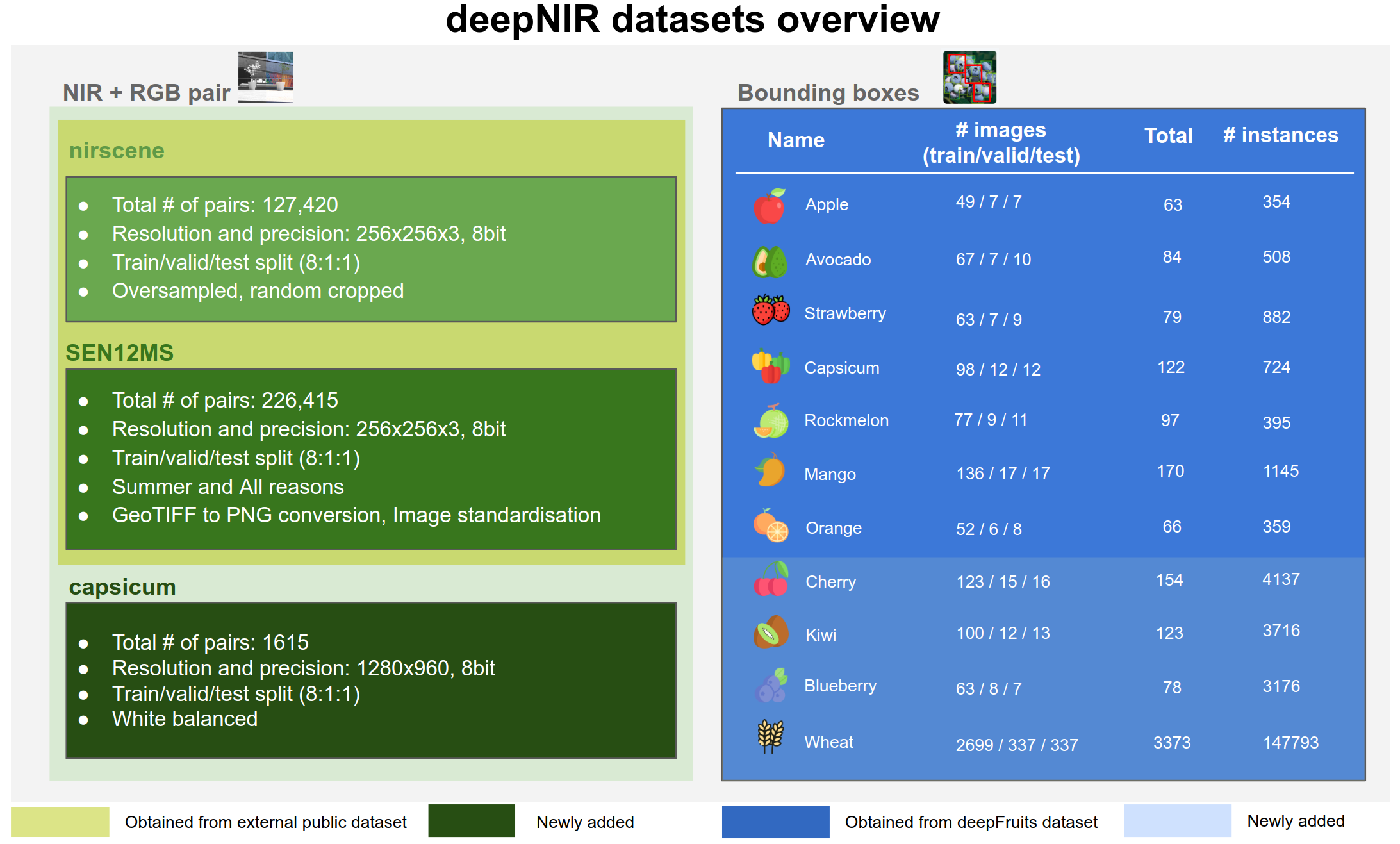}
\caption{NIR+RGB pair and object detection datasets overview}
\label{fig:dataset-overview}
\end{figure}

This paper is different to \cite{An2019-xl, Yuan2020-ma} in the following aspects. First and most importantly, we clearly present experimental results with a high level of technical detail which are lacking in both. In synthetic image generation and generally machine learning, it is one of important tasks to carefully split train and test sets to hold equivalent or similar statistical properties (e.g., feature distributions) among sets. But none of them correctly disclosed this, only presented scores which is somehow meaningless. None of them made the dataset available for public use so that it is impossible to reproduce or to build another system on top of their studies. Lastly, it is vague why and how synthetically generated information is useful in their work. We demonstrate this by feeding forward synthetic NIR images into a subsequent fruit detection task. Therefore our contributions in this paper are:
\begin{itemize}
\item We made publicly available NIR+RGB datasets for synthetic image generation\cite{sa2022deepnir}\footnote{Note that we utilised 2 publicly available datasets; nirscene\cite{Brown2011-hg} and SEN12MS\cite{Schmitt2019-oo} in generating these synthetic datasets. For image pre-processing, oversampling with hard random cropping was applied to nirscene and image standardisation (i.e., $\mu=0$, $\sigma=1$ was applied. It is first time to realise capsicum NIR+RGB dataset. All datasets are split as 8:1:1 (train/validation/test) ratio}. This dataset follows a standard format so that it is straight forward to be exploited with any other synthetic image generating engine.
\item Expanded our previous study\cite{Sa2016-zv}, and we added 4 more fruit categories including their bounding box annotations. 11 fruit/crops were rigorously evaluated and to our best knowledge this is the largest type of dataset currently available. 
\end{itemize}
Figure \ref{fig:dataset-overview} summarises all datasets presented in this paper. The left green boxes indicate NIR + RGB pair datasets and their technical detail and the right blue box denotes bounding box dataset for object detection.

Other than above contributions, we also present detailed experimental results, their analysis and insights that can be useful for readers. The dataset can be downloaded from

\centering{\url{http://tiny.one/deepNIR}}

\justifying

The rest of paper is structured as follow. Section \ref{sec:related work} presents literature reviews on synthetic image generation and object detection. Section \ref{sec:method} covers methodologies used for generating synthetic images, and dataset detail, and the concise summary of the generative adversarial network and a single stage detection framework. Section \ref{sec:results} contains evaluation metrics for both synthetic image generation and object detection followed by qualitative and quantitative results. This also includes inter-comparisons between models developed in this paper and baseline comparisons with evaluations from other studies. We also discuss advantages and limitations that we found in section \ref{sec:discussion}. Section \ref{sec:conclusions} concludes the paper by giving a summary of results, impact of the proposed work, and future outlook.

\section{Related work}\label{sec:related work}
In this section, we describe previous related studies; especially focusing on public RGB-NIR datasets, synthetic image generation, and object detection approaches. This section presents clear distinction between other datasets and what we proposed and state-of-the-art synthetic image generation and object detection techniques that other researchers can alternatively exploit. Therefore this section is helpful for readers to better understand the remaining sections.

\subsection{Near-infrared(NIR) and RGB image dataset}

Modern data-driven deep learning approaches have demonstrated very promising and impressive performance in a variety of sectors, and it is no exaggeration to say that large-scale and high-quality training datasets have played a pivotal role in achieving these successes. Especially within the agricultural domain, NIR-RGB (or multispectral interchangeably) datasets provide rich feature information about vegetation such as crops or fruits. They are regarded as one of the important key indices for agronomists, data scientists and machine learning researchers. Hence there exist valuable and noticeable contributions \cite{Chebrolu2017-up,Sa2018-fu,Sa2018-sg,Sa2015-pl,Di_Cicco2016-rx} in broad-arcs, horticulture\cite{Sa2017-ge}, or protected farm scenarios\cite{Lehnert2016-pl, Lehnert2020-kx}.

Brown M. \textit{et al.}\cite{Brown2011-hg} is one of frontier NIR-RGB datasets focusing on scene recognition tasks by utilising the multispectral information. As mentioned earlier, this dataset contains 477 RGB+NIR images pairs that were asynchronously captured using two cameras in mostly outdoor daily-life scenes. This dataset is useful, and we also utilised it in this paper. Still, a temporal discrepancy in a pair and lack of radiometric calibration and small scale dataset are challenges of the use of this dataset. We demonstrate the impact of dataset scale and oversampling strategy in the following section.

More recently, one of our previous studies\cite{Sa2015-pl,McCool2016-sk} that focused on sweet pepper detection and semantic segmentation using multispectral images contributed to the horticulture sector. 103 pixel-level annotations and NIR+RGB pairs were used in this work. Considering challenges in the agricultural scene pixel-labelling task, it was one of the novel datasets together with \cite{Haug2015-aw} that contributed 60 annotations even though the scale was relatively small than nowadays datasets. More importantly, we share in this paper another 1615 NIR-RGB pair dataset collected in that campaign yet annotated.

Chebrolu N. \textit{et al.}\cite{Chebrolu2017-up} offered a comprehensive large-scale agricultural robot dataset that is suitable for vegetation semantic segmentation as well as localization and mapping. Multispectral images, RGB-D, LiDAR, RTK-GPS, and wheel odometry sensor data were collected over a sugar beet field for two months in Germany. 5TB of such dataset was obtained, but \cite{Chebrolu2017-up} does not provide a dataset summary table, so it is rather difficult to find how many multispectral images and their annotations without attempting to use the dataset.

In agriculture, satellite imagery is one of important resources so that is widely utilised in many applications key source information. They provide abundant, large scale earth observations, which are useful for data-driven machine learning approaches. Therefore, promising studies \cite{Segarra2020-xt} and a public dataset\cite{Schmitt2019-oo} utilised satellite multispectral imagery (e.g., Sentinel-2 A+B twin satellites platforms launched by the European Space Agency (ESA)).

Schmitt M. \textit{et al.}\cite{Schmitt2019-oo} introduced an unprecedented multi-spectral dataset in 2019. They sampled from 256 globally distributed locations over four seasons which constitutes about 180\unit{k} NIR+RGB pairs. We adopt this dataset with the following processes in this paper. We converted the raw dataset formatted as multi-channel GeoTIFF into an ordinary image formation with image standardisation for each image pair and split it to train/validation/test sets. We agree that these steps are trivial and straightforward to achieve. However, in practice especially training a model using 180\unit{k} multispectral images, it often matters to know concise and exact split sets and to secure a direct trainable dataset rather than an ambiguous and questionable split\cite{Yuan2020-ma}\cite{An2019-xl} or requiring any modifications from a reproducibility perspective. This ambiguity can result in spending time and effort building up a baseline. For instance, converting a digital number (DN) to 8bit per channel standard image format can be a non-trivial task, and evaluation metrics can also vary depending on how the dataset is split. From these perspectives, it is important to establish a fixed dataset split with the corresponding metric to see how much other factors impact the performance (i.e., ablation studies).

\begin{table}
\caption{Summary of related studies on NIR-RGB dataset} \label{tbl:nir-rgb-dataset}
\begin{center}
\begin{tabular}{ccc}
\textbf{Study}                                                                               & \textbf{Advantages}                                                                                        & \textbf{Can improve}                                                                                                                                            \\
\hline
Brown M. \textit{et al.}\cite{Brown2011-hg}       & \begin{tabular}[c]{@{}l@{}}Outdoor \\ daily-life scenes\end{tabular}                        & \begin{tabular}[c]{@{}l@{}}- A temporal discrepancy in a pair\\ - Lack of radiometric calibration, \\ - Small scale dataset to use\end{tabular} \\
\hline
Chebrolu N. \textit{et al.}\cite{Chebrolu2017-up} & Comprehensive large-scale dataset                                                           & \begin{tabular}[c]{@{}l@{}}lack of comprehensive \\ dataset summary table\end{tabular}                                                          \\
\hline
Schmitt M. \textit{et al.}\cite{Schmitt2019-oo}   & \begin{tabular}[c]{@{}l@{}}Sampling from different locations \\ over 4 seasons\end{tabular} & Requires preprocessing to use                                                                                                                 
\end{tabular}
\end{center}
\end{table}

\subsection{Synthetic image generation}

Synthetic image generation is perhaps one of the most attractive and active fields among many other interesting and promising applications of deep learning techniques. Rooting from Generative Adversarial Networks (GAN)\cite{Goodfellow2014-by}, there are brilliant ideas that either improved the original adversarial idea or established another stepping stone.

Mirza M. \textit{et al.}\cite{Mirza2014-rl} proposed a new idea that conditionally takes not only noise input for a generator but auxiliary information for better model convergence and generalisation (cGAN). They demonstrated the impact for the image-to-image translation task and enabled/highly influenced other variants such as Pix2pix\cite{Isola2017-hr,wang2018pix2pixHD}, StyleGAN\cite{Karras2020ada}, CycleGAN\cite{CycleGAN2017}, or more recently OASIS\cite{onfeld2021you}. Even though they propose different approaches and applications, the fundamental idea stems from the studies mentioned above. In this paper, we adopt the work of \cite{wang2018pix2pixHD} in order to evaluate and confirm the implications of our dataset, but readers can freely choose any state-of-the-art framework as a synthetic generation tool.

On top of these core works, many interesting applications use synthetic image generation. Transforming from visible range (RGB) to near-infrared spectrum(NIR) is demonstrated in \cite{An2019-xl}, and \cite{Yuan2020-ma} using nirscene and SEN12MS datasets respectively. Aslahishahri M. \textit{et al.}\cite{Aslahishahri2021-op} showed aerial crop monitoring with synthetic NIR generation using a software package from \cite{Isola2017-hr}. Although they disclosed their dataset, its scale (only 12 pairs) is too marginal to use for model training effectively.

There were interesting studies that transformed from thermal range to visual spectrum \cite{Bhat2020-fg}\cite{Berg2018-wr}. The objective was the estimation of nonlinear mapping between visible spectrum (450$\sim$750\unit{nm}) to long wave infrared (8$\sim$12\unit{um}) range. The large gap in the spectrum causes severe appearance difference, hence the task is more difficult than the NIR-RGB mapping case. In order to achieve a good generalised model and stable performance, a large scale dataset and precise thermal calibration (e.g., fluid field correction and temperature calibration) are required.

Instead of separately treating each image, there were attempts to fuse only distinct features. The studies of \cite{Li2021-jp, Ma2019-fa} proposed a fusion approach of multimodal data and the goal of their work was to fuse visible texture from an RGB image and thermal radiation from an infrared image by forcing a discriminator to have more details.

As we can see from the above literature, it is key to choose proximal spectrum ranges to learn a nonlinear mapping successfully. Ma .Z \textit{et al.}\cite{Ma2021-gv} exemplified this by demonstrating transformation from NIR-I (900$\sim$1300\unit{nm}) to NIR-IIb (1500$\sim$1700\unit{nm}) in vivo fluorescence imaging\footnote{An imaging technique applying glow substances to cells to record responses of live organisms.}. According to their results, they achieved unprecedented signal-to-background ratio and light-sheet microscopy resolution. A similar approach was applied to medical imagery\cite{Welander2018-rb}; generating magnetic resonance images (MRI) from computed tomography image (CTI) using CycleGAN\cite{CycleGAN2017}and unsupervised image-to-image translation network(UNIT)\cite{Liu2017-qd}.

We have introduced the most fundamental studies and outstanding applications in our perspectives in image-to-image translation using GAN techniques. However this research field is active and developing at a fast pace, so we would like to refer to a more solid and recent survey paper\cite{Soni2021-tg}\cite{Fawakherji2021-jc}.

\subsection{Object-based fruit localisation}
Synthetically generated images from the previous section can be used as auxiliary information for various computer vision tasks such as object classification, recognition, bounding box-level detection, and semantic segmentation in order to improve performance. In this paper, we are interested in the fruit object detection (i.e., bounding box localisation) task following our previous studies presented in DeepFruits\cite{Sa2016-zv} where we demonstrated 7 fruits/crops detection using a two-stage object detector\cite{Ren2015-no}. On top of the work, we share 4 additional novel fruit annotations and their split. We evaluated our 3 and 4 channel datasets using a single-stage detector\cite{glenn-jocher-2021-5563715}.

The object detection problem is one of the most important tasks in remote sensing, computer vision, machine learning, and robotics communities. Recent advances in large scale datasets and machine learning algorithms accelerated by GPU computing have unlocked potential and achieved super human-level performance. In this research area, there are two main streams; single and two-stage detection. 

The first formulates the problem as a single regression optimisation problem gaining faster inference speed with the cost of inferior performance\cite{glenn-jocher-2021-5563715,Ge2021-un}. Whereas the latter, two-stage detector, employs a region proposal network(RPN), which suggests a number of candidates (e.g., rectangles or circle\cite{Yang2020-id} primitives) for the subsequent object classification and bounding-box regression heads\cite{Ren2015-no,He2017-pf,Carion2020-lx}. Generally, this approach achieved superior performance at the cost of processing speed. According to the recent trends in object detection, it is worth mentioning that the detection performance gap between these paradigms has been significantly reduced with the aid of intensive optimised image augmentation techniques and more efficient network architecture design\cite{mmdetection}. We will discuss this more in section \ref{subsec:fruit-detection}.

Even though we presented the most remarkable achievements in the area, we would like to point out another object detection survey paper\cite{Liu2020-zy} to cover more concrete summaries and research directions.

\section{Methodologies}\label{sec:method}
In this section, we present synthetic near-infrared image generation and its application, object detection, by using the generated 4 channels (i.e., 3 * visible + 1 * infrared spectra) of data.

\subsection{Synthetic near-infrared image generation}

Abundant and high-quality training data is one of the essential driving factors, especially for data-driven approaches such as deep neural networks (DNN). Securing such data often requires a lot of resources (e.g., manual annotations). Therefore, researchers and communities have been devoted tremendous efforts to this, which leads to impressive and brilliant ideas. Data augmentation \cite{info11020125,Shorten2019-ne}, Pseudo labeling \cite{Xie2019-ue}, and generative adversarial models \cite{Isola2017-hr}\cite{wang2018pix2pixHD}\cite{onfeld2021you}. In this paper, we are interested in exploiting a generative model for the following reasons. First, it is straightforward to re-formulate the problem by adopting ideas from previous studies such as style-transfer\cite{Isola2017-hr} and fake image generation\cite{Karras2021-xk}. For the training phase, we only need to feed image pairs (RGB, NIR) as input and target. Second, there exist well-established resources that demonstrate outstanding performance in non-agricultural domains such as fake face image generation or style transfer from hand drawings to masterpieces.

\begin{figure}
\centering
\includegraphics[width=\textwidth]{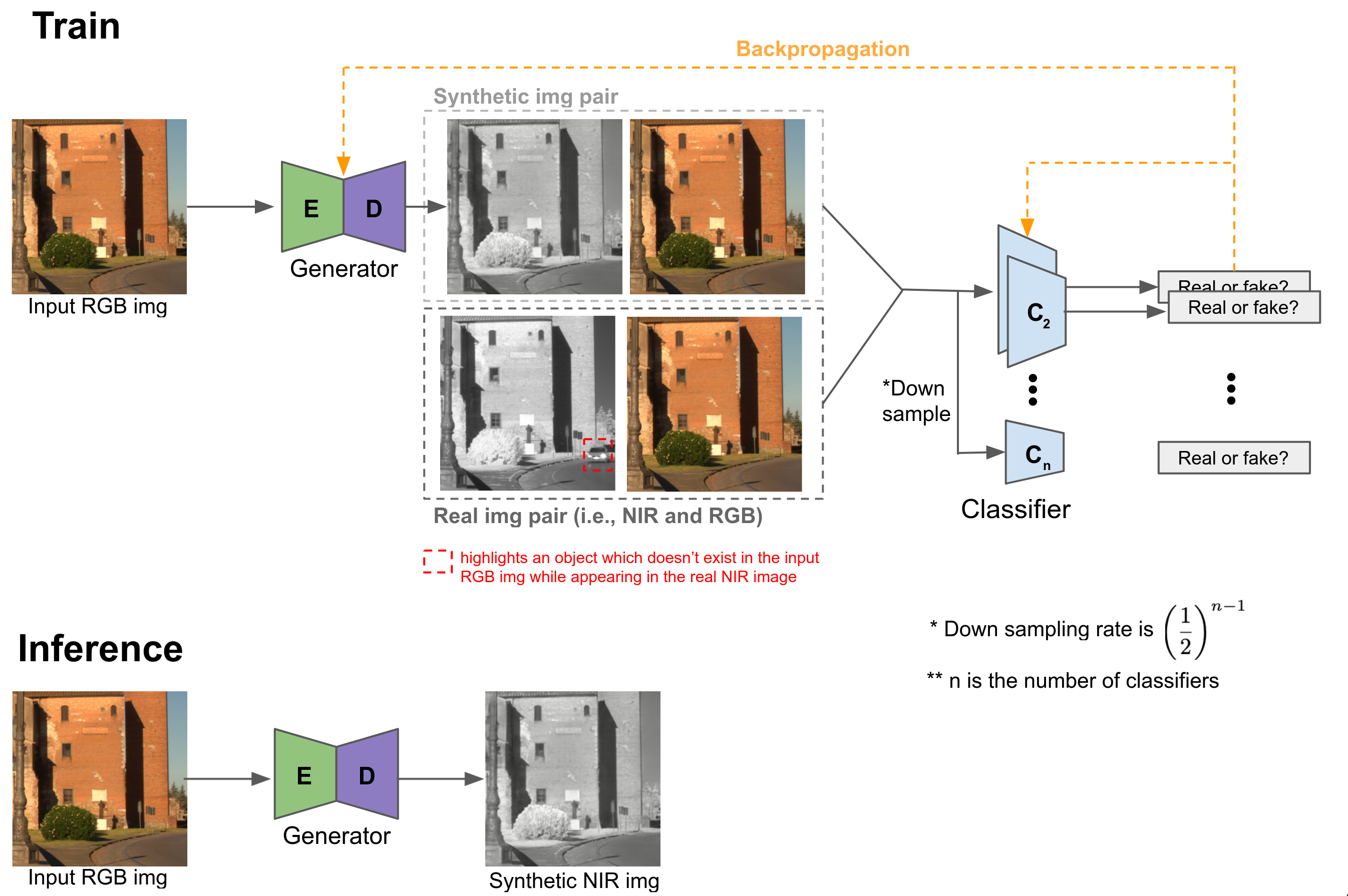}
\caption{Synthetic image train (top) and inference (bottom) pipeline. This is one of typical type of conditional Generative Adversarial Network (cGAN)\cite{Isola2017-hr}. Image redrawn from \cite{wang2018pix2pixHD}.}
\label{fig:pix2pi2}
\end{figure}

Figure \ref{fig:pix2pi2} illustrates one of the generative adversarial networks (GAN) for synthetic image generation \cite{wang2018pix2pixHD}. Our goal is to find the optimal generator and classifier given real image pairs at the training phase. As shown in the figure, the role of the generator and classifier is creating a synthetic image pair and distinguishing real or synthetic pairs, respectively. The inference stage simply performs forward prediction using the trained generator model and input RGB image, creating synthetic image output. It is worth mentioning that the generator may have abilities to generate small obscured scenes if train and test datsaets share a similar context. For instance, there is a passing car in the real NIR image but not in the RGB image in Figure \ref{fig:pix2pi2}. This happened because NIR and RGB images were asynchronously captured in the public dataset \cite{Brown2011-hg}. Despite this, our generator is able to recover the small portion of image (red dashed box) because it learnt how to transfer from RGB spectrum (380\unit{nm}-740\unit{nm}) to NIR ($\sim$ 750\unit{nm}).

More formally, the objective of GAN (conditional GAN\cite{Mirza2014-rl} more precisely) can be expressed:

\begin{equation}
\mathcal{L}_{\mbox{\tiny{GAN}}}(G_{\theta_{G}}, D_{\theta_{D})} = \mathbb{E}_{x\sim p_{data}(x), y}[logD_{\theta_{D}}(x|y)] + \mathbb{E}_{z\sim p_{z}(z)}[log(1-D_{\theta_{D}}(G_{\theta_{G}}(z|y))]
\end{equation}
where $G_{\theta_{G}}$ and $D_{\theta_{D})}$ are generator, $G : \{x, z\} \mapsto y$ and classifier (or discriminator) parameterised $\theta_{G}$ and $\theta_{D}$ respectively. $x\in\mathbb{R}^{\text{WxHxC}}$, $y\in\mathbb{R}^{\text{WxHxC}}$ and $z\in\mathbb{R}^1$ are input image, target image, and a random noise in this case. Intuitively the first term indicates the expectation of classifier given data sample $x$ (i.e., RGB image) drawn from input data distribution and target $y$, which is an NIR image. Maximising this term implies we successfully fool the classifier even though the generator produces synthetic images. The second term is what we want to minimise the difference between the output of generator($\in\mathbb{R}^{\text{WxHxC}}$) given random noise $z$ drawn from noise distribution, $p_z(z)$, given target image $y$ and target image $y$ as close as possible.

Concretely we can also add $\mathcal{L}1$ loss function in order to minimise blurring

\begin{equation}
\mathcal{L}_{L1}(G_{\theta_{G}}) = \mathbb{E}_{x,y,z}[||z - G_{\theta_{G}}(x|y)||_{1}]
\end{equation}

Therefore, the final objective is a min-max optimisation problem
\begin{equation}
arg \operatorname*{min}_{G_{\theta_{G}}} \operatorname*{max}_{D_{\theta_{D}}}  \mathcal{L}_{\mbox{\tiny{GAN}}}(G_{\theta_{G}}, D_{\theta_{D})} + \lambda \mathcal{L}_{L1}(G_{\theta_{G}}).
\end{equation}

Other than conditional GAN, there are many GAN variants in formulating loss functions such as Convolutional GAN\cite{Radford2015-au} or Cycle GAN\cite{Zhu2017-qj} which also can be utilised for synthetic image generation tasks.

Among many possible approaches, we selected the Pix2pixHD framework \cite{wang2018pix2pixHD} as our baseline study for the following reasons. It has been widely used among synthetic data generation tasks, so there are many comparable resources. It can handle higher resolution images than its ancestor\cite{Isola2017-hr} and is easy to use with many available options such as hyperparameter searching and model evaluation. We made datasets used in this paper for training and testing available for the public. One can reproduce or evaluate model performance using different state-of-the-art GAN frameworks.

\subsubsection{Datasets used for generating synthetic image}

We made minor modifications in the use of our baseline synthetic image generation framework (i.e., Pix2pixHD) to be able to evaluate model performance while varying datasets, as shown in Table \ref{tbl:synthetic-dataset}. With these datasets, one will be able to reproduce the similar results we achieved or use other frameworks for superior outcomes.

\begin{table}
\caption{Summary of datasets used for generating synthetic image} \label{tbl:synthetic-dataset}
\begin{center}
\begin{tabular}{cccccccccc}
\textbf{Name}              & \textbf{Desc.}                                         & \textbf{\# Train} & \textbf{\# Valid} & \textbf{\# Test} & \textbf{Total} & \textbf{\begin{tabular}[c]{@{}c@{}}img size \\ (wxh)\end{tabular}}& \textbf{\begin{tabular}[c]{@{}c@{}}Random \\ cropping\end{tabular}} & \textbf{\begin{tabular}[c]{@{}c@{}}Over\\ sampling\end{tabular}} & \textbf{\begin{tabular}[c]{@{}l@{}}Spectral\\ range(nm)\end{tabular}} \\ \hline
\multirow{5}{*}{nirscene1} &                                                        & 2880              & 320               & 320              & 3520           & 256x256                & Yes                                                                 & x10                                                              & \multirow{5}{*}{\begin{tabular}[c]{@{}l@{}}400-\\ 850\end{tabular}}   \\
                           &                                                        & 14400             & 1600              & 1700             & 17700          & 256x256                & Yes                                                                 & x100                                                             &                                                                       \\
                           &                                                        & 28800             & 3200              & 3400             & 35400          & 256x256                & Yes                                                                 & x200                                                             &                                                                       \\
                           &                                                        & 57600             & 6400              & 6800             & 70800          & 256x256                & Yes                                                                 & x400                                                             &                                             
                           \\ \cline{2-10} 
\multirow{2}{*}{SEN12MS} & \begin{tabular}[c]{@{}c@{}}All \\ seasons\end{tabular} & 144528            & 18067             & 18067            &      180662          & 256x256                & No                                                                 & N/A                                                              & \multirow{2}{*}{\begin{tabular}[c]{@{}l@{}}450-\\ 842\end{tabular}}   \\
                           & Summer                                                 & 36601             & 4576              & 4576             &     45753           & 256x256                & No                                                                 & N/A                                                              &                                                                       \\ \cline{2-10} 
capsicum                   &                                                        & 1291              & 162               & 162              & 1615           & 1280x960               & No                                                                  & N/A                                                              &     \multirow{2}{*}{\begin{tabular}[c]{@{}l@{}}400-\\ 790\end{tabular}} \\                       
\end{tabular}
\end{center}
\end{table}

\begin{figure}
\centering
\includegraphics[width=\textwidth]{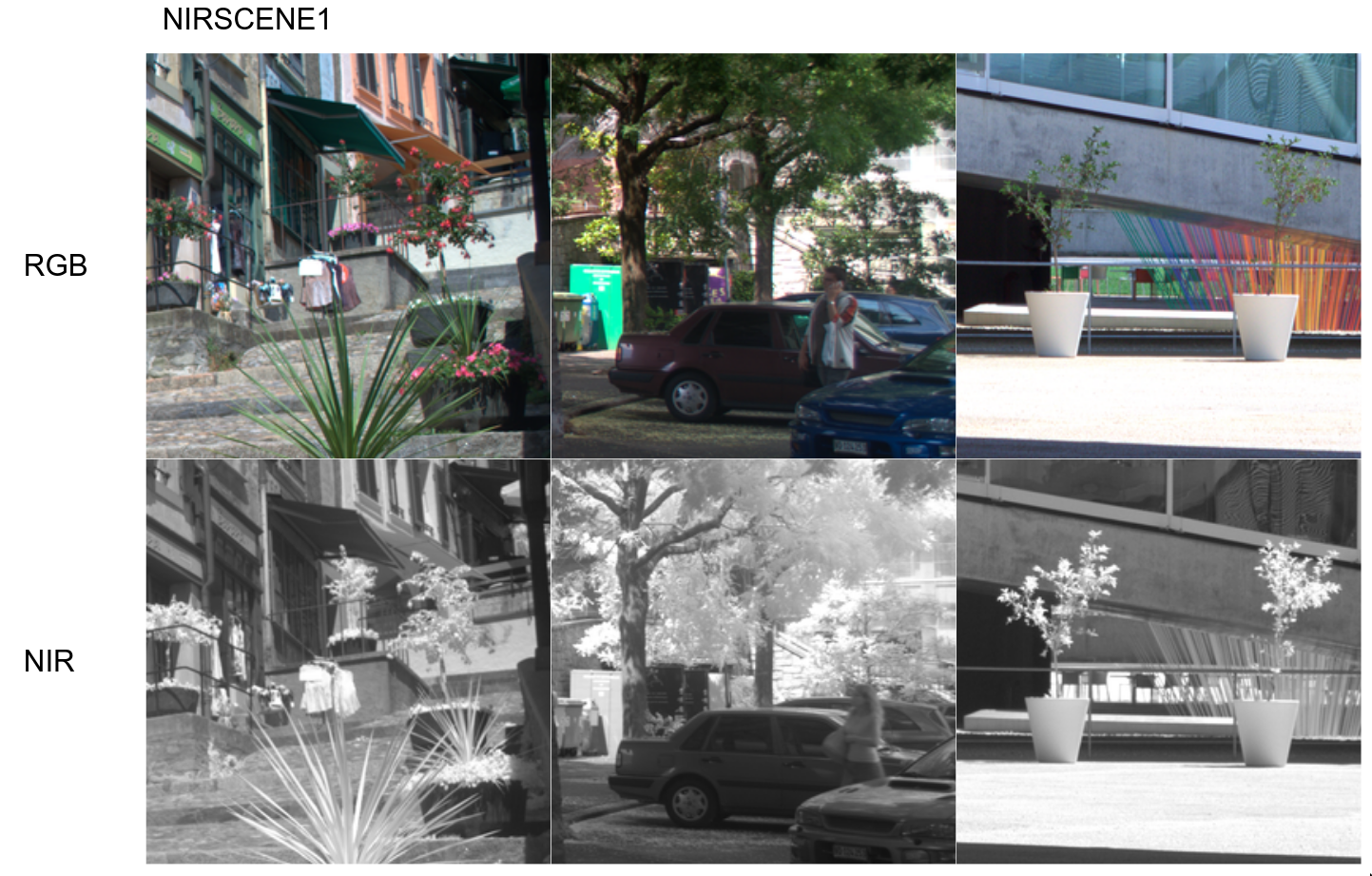}
\caption{RGB (top row)+NIR (bottom row) samples drawn from our nirscene1 dataset (256$\times$256). This dataset was mostly taken in outdoor environments with band-cutoff filter equipped two digital single-lens reflex(DSLR) cameras. Note that there also exists noticeable temporal gap between NIR and RGB images (see pedestrians in the middle column).}
\label{fig:nirscene1-sample}
\end{figure}

The data consist of 3 public \cite{Brown2011-hg}\cite{Schmitt2019-oo}\cite{Sa2015-pl} datasets. The first dataset namely, \textit{nirscene1}, contains 477 RGB+NIR images (1024x679) that were captured with commercial high-end cameras with a 750\unit{nm} band-cutoff filter. The colour images were white-balanced, and the channel-wise average was applied to the infrared images. Two image alignment (or registration) was done through feature matching in RGB and NIR domains. These are only critical characteristics, but more technical detail can be found from \cite{Brown2011-hg}. Figure \ref{fig:nirscene1-sample} shows sample images from the dataset. This dataset is useful but only has 477 pairs which may hinder a good visible-infrared domain mapping. Although more experimental results will be presented in the following section \ref{sec:results}, we performed hard-cropping and over-sampling to resolve this issue. Hard-cropping is one of the augmentation techniques which generates cropped datasets, whereas soft-cropping generates cropped samples during the training/testing phase. Over-sampling refers to randomly sampling more redundant data. It is a fact that the maximum amount of information we can get from the over-sampling is the original data. However, we found that over-sampling helped stabilise training and improved performance by a large margin. Regarding this, we will discuss and analyse more in the result section \ref{sec:results}.

\begin{figure}
\centering
\includegraphics[width=\textwidth]{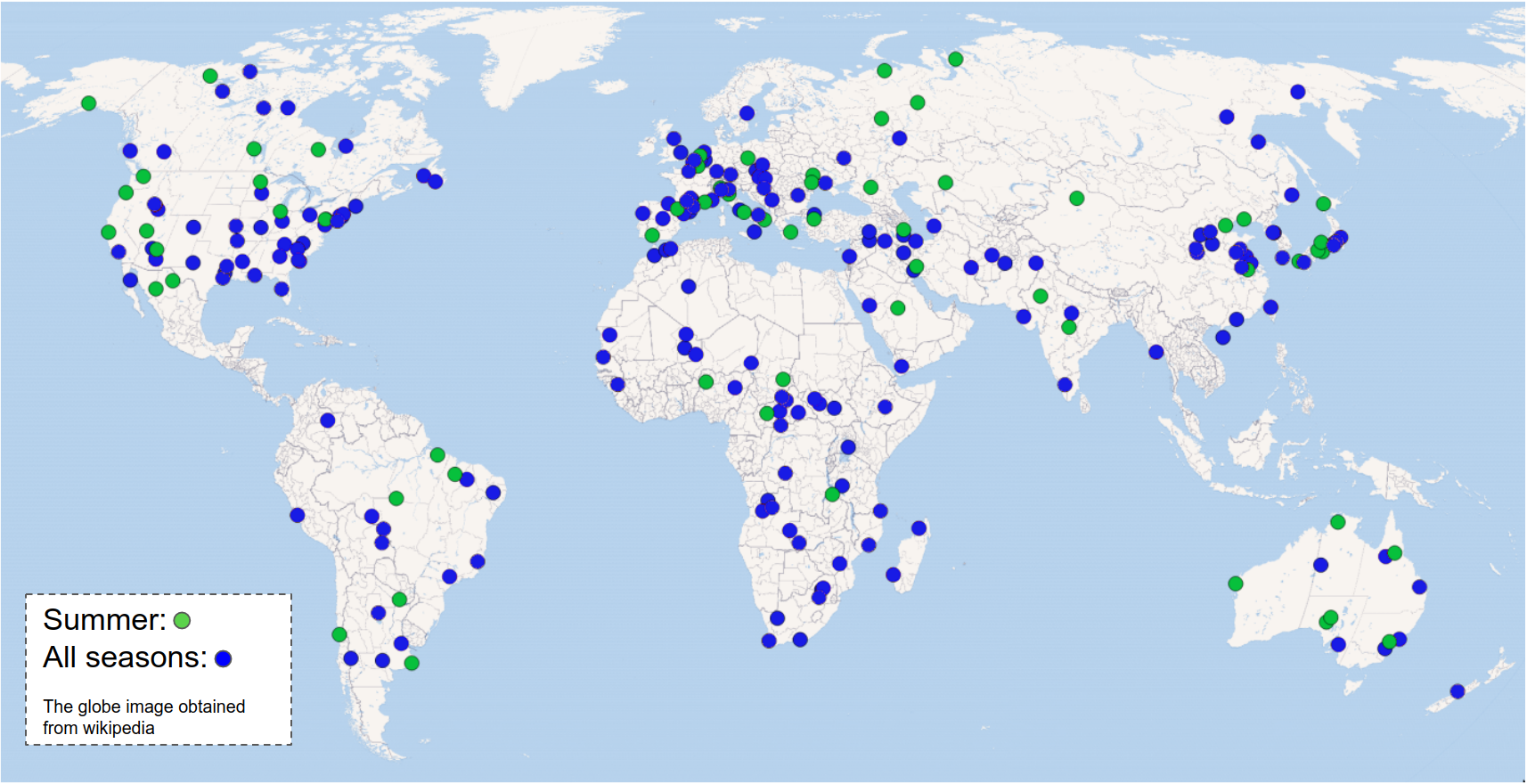}
\caption{Green indicates 65 sampled locations from SEN12MS summer dataset and blue covers all seasons and 256 globally distributed locations in total.}
\label{fig:sen12ms-www}
\end{figure}

\begin{figure}
\centering
\includegraphics[width=\textwidth]{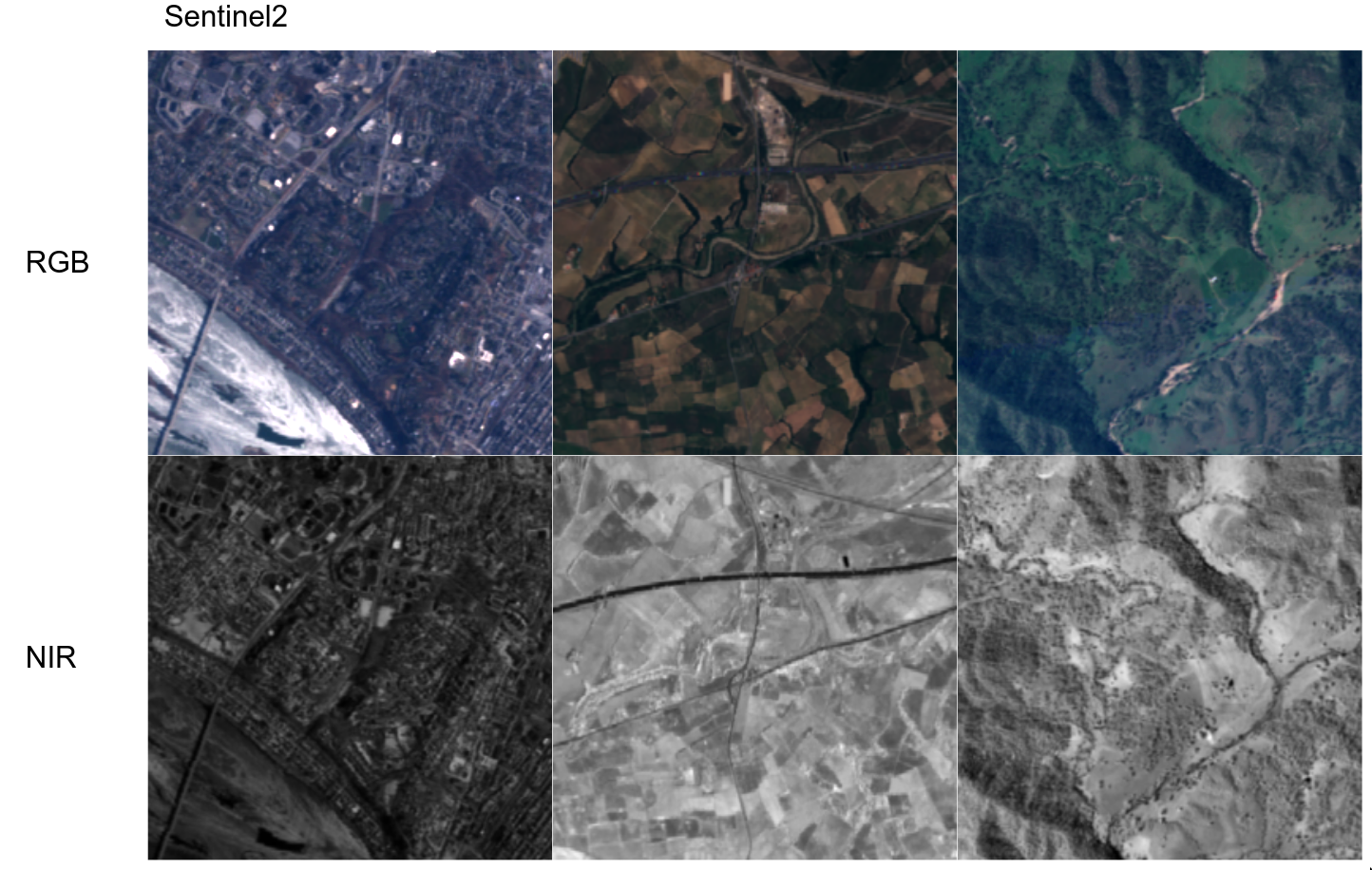}
\caption{RGB (top row) and NIR (bottom row) samples from our SEN12MS dataset (random samples in winter and spring). Note that left RGB image's brightness is adjusted for better visualisation. Original RGB image is darker.}
\label{fig:sentinel2-sample}
\end{figure}

The second dataset, \textit{SEN12MS}, is publicly available satellite imagery from Sentinel-1 and Sentinel-2. No image processing is applied to this dataset, we only selected two subsets (Summer and All seasons) followed by train/valid/test split. Figure \ref{fig:sen12ms-www} shows geolocations where the authors sampled multispectral imagery. A multispectral image covers spectral range from 450\unit{nm} - 842\unit{nm} (i.e., band2, band3, band4, band8 of Sentinel-2) and captured at 768\unit{km}. This leads to having 10\unit{m} ground sample distance (GSD)/pixel. Radiometric calibration was properly performed by the satellite system organisation. Figure \ref{fig:sentinel2-sample} demonstrates a couple of samples in this dataset.

\begin{figure}
\centering
\includegraphics[width=\textwidth]{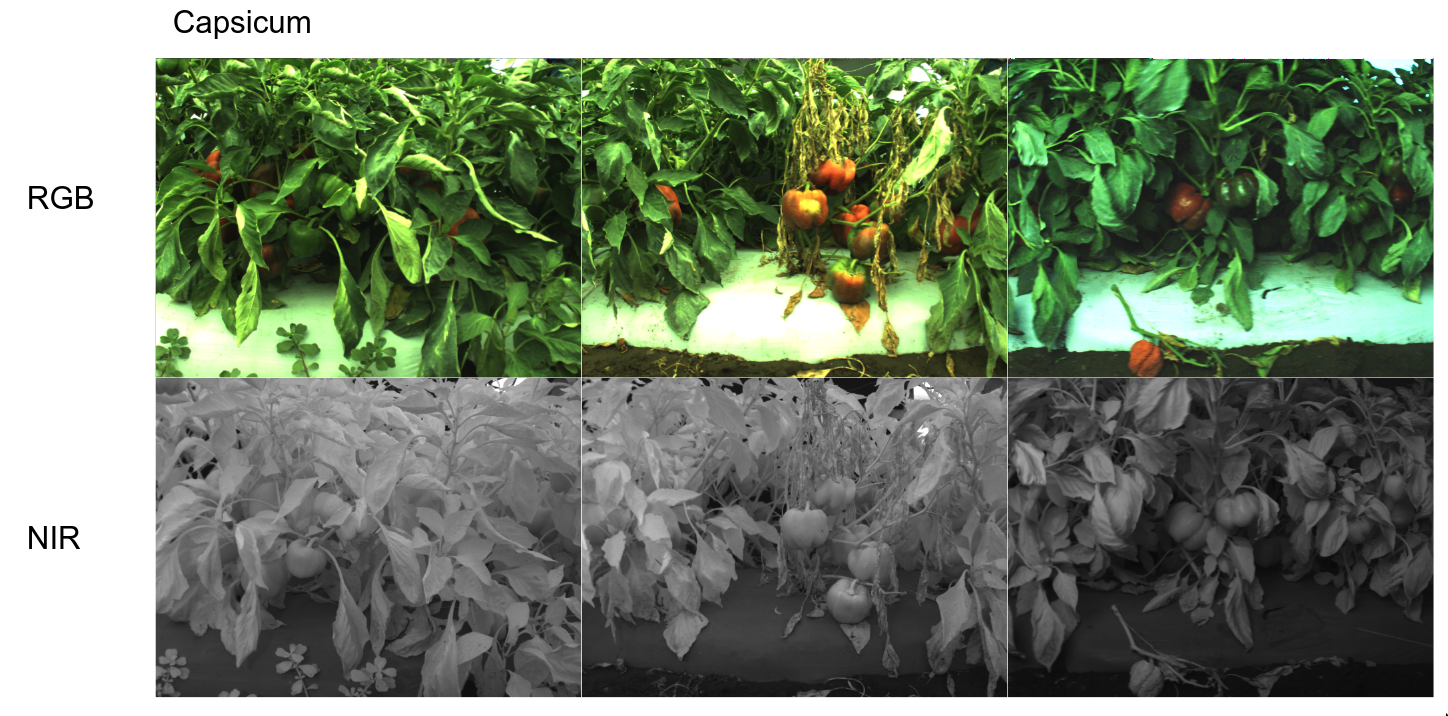}
\caption{RGB (top row) and NIR (bottom row) samples from our capsicum dataset. It is one of must challenging datasets containing very cluttered and complex scenes in close proximity. Note that right RGB image's brightness is adjusted for better visualisation. The original RGB image is darker.}
\label{fig:capsicum-sample}
\end{figure}

The last dataset, \textit{capsicum}, is one of our previous studies presented in \cite{Sa2015-pl}\cite{Sa2016-zv}. The dataset was collected from sweet pepper farms in Australia, Gatton and Stanthorpe, with a multispectral camera, JAI AD-130GE. This camera has two charges coupled device (1280x960) prism mechanisms for each RGB and NIR spectrum. Unlike other datasets, we use the larger original image to train our model because this simplifies the subsequent procedures (e.g., object detection). Data collection campaigns were mostly performed at night with controlled visible-, infrared light sources to mitigate external interference. White balance was properly performed with a grey chart, and radiometric calibration was omitted. Figure \ref{fig:capsicum-sample} shows samples from this dataset.  

\subsection{Fruit detection using synthetic images}\label{subsec:fruit-detection}
In this section, we present one of the applications for which we can take advantage of synthetically generated images from the previous step. Object detection or semantic image segmentation is one of the important downstream tasks in many research and commercial areas. Especially, precise object detection in agriculture can be considered as a pivotal stepping stone because it can be used for many other subsequent tasks such as crop counting, yield estimation, harvesting and disease detection with bounding-box level segmentation or its classification.

For objection detection, we chose Yolov5\cite{glenn-jocher-2021-5563715} mainly due to fast inference time (i.e., single-stage detection), easy-to-train and intuitive visualisation advantages over other frameworks. However, there exists other powerful frameworks such SAHI\cite{akyon2022sahi}, Detectron2\cite{wu2019detectron2}, or MMDetection\cite{mmdetection} that can also be exploited. These frameworks are very flexible in adapting new modules or datasets, and supports many pretrained weights which can improve object detection performance by large margin.

The network architecture and implementation details of Yolov5 can be found from \cite{glenn-jocher-2021-5563715}, and we present only a concise high-level view in order to help readers in understanding object detection.

It consists of four sub-parts; namely input, backbone, neck, and head layers. The first input layer adopts mosaic data augmentation that is an aggregation of cropped images, adaptive anchor, and many other augmentation techniques\cite{info11020125}. Backbone and neck networks are in charge of feature extraction by making use of focus (i.e., image slicing), convolution-Batch-Normalisation, and Leaky ReLU (CBL), Cross-Stage-Partial(CSP)\cite{Wang2019-bg}, and Spatial Pyramid Pooling (SPP)\cite{He2015-gf}, Feature Pyramid Networks (FPN)\cite{Lin2016-nm} and Path Aggregation Network (PAN)\cite{Liu2018-ob} modules. Intuitively, the output of the neck network is feature pyramids that incorporate varying object scales, which may lead to superior performance than other single-stage-detectors (SSD). The last head layer is an application-specific layer, and most object detection tasks predict bounding box (4), confidence (4), and class(1) from the head layer. More concretely, bounding box loss, $\mathcal{L}_{\mbox{\tiny{box}}}$ used in the object detection is expressed

\begin{align}
    \mathcal{L}_{\mbox{\tiny{box}}} =& \sum_{i=0}^{s\times s}\sum_{j=0}^{N}I_{i,j}^{\mbox{\tiny{obj}}}(1 - \text{GIoU}) \\
    I_{i,j}^{\mbox{\tiny{obj}}} =& \begin{cases}
    1 & \text{if prediction exists within annotation}\\
    0 & \text{otherwise.}
    \end{cases}\\
    \text{where GIoU} =& \text{IoU} - \frac{C - \mathcal{U}}{{C}}\\ \text{IoU}=&\frac{\mathcal{I}}{\mathcal{U}},\;\;\;\mathcal{U} = \hat{B} + B - \mathcal{I}
\end{align}
where $s$ is the number of a grid, $N$ is the number of bounding boxes in each grid. GIoU is generalised intersection over union\cite{Rezatofighi2019-gw} which has [-1, 1] scalar value. $\hat{B}$, $B$ are area of prediction and annotation bounding boxes (i.e., two arbitrary convex shapes) and $C$ is the area of the smallest enclosing convex shape. $\mathcal{I}$ and $\mathcal{U}$ are the intersection and union of $\hat{B}$ and $B$ respectively. IoU indicates intersection over union of $\hat{B}$ and $B$. Intuition of the loss function is that the loss will keep increasing with smaller GIoU implying smaller overlap between $\hat{B}$ and $B$, on the other hand the loss decreases with larger GIoU when two bounding boxes are largely overlapped.

There are two more losses for confidence score, $\mathcal{L}_{\mbox{\tiny{score}}}$ and class probability $\mathcal{L}_{\mbox{\tiny{class}}}$ and they are modeled by logistic regression and binary cross entropy as follow.

\begin{gather*}
    \mathcal{L}_{\mbox{\tiny{score}}} = -\sum_{i=0}^{s\times s}\sum_{j=0}^{N}I_{i,j}^{\mbox{\tiny{obj}}}\bigg(S_{i}^{j}\text{log}(\hat{S}_{i}^{j})+(1-S_{i}^{j})\text{log}(1-\hat{S}_{i}^{j})\bigg) \\ \quad\quad-\lambda_{\mbox{\tiny{empty}}}\sum_{i=0}^{s\times s}\sum_{j=0}^{N}I_{i,j}^{\mbox{\tiny{empty}}}\bigg(S_{i}^{j}\text{log}(\hat{S}_{i}^{j})+(1-S_{i}^{j})\text{log}(1-\hat{S}_{i}^{j})\bigg)\\
I_{i,j}^{\mbox{\tiny{empty}}} = \begin{cases}
    1 & \text{if there is no prediction}\\
    0 & \text{otherwise.}
    \end{cases}
\end{gather*}

where $\hat{S}_{i}^{j}$ and $S_{i}^{j}$ indicate the prediction and annotation scores (usually = 1.0) of the $j$-th bounding box in the $i$-th grid. $\lambda_{\mbox{\tiny{empty}}}$ is the weight when there is no predicted object within the $j$-th bounding box.

Similarly, class probability $\mathcal{L}_{\mbox{\tiny{cls}}}$ is defined as
\begin{gather}
    \mathcal{L}_{\mbox{\tiny{cls}}} = -\sum_{i=0}^{s\times s}\sum_{j=0}^{N}I_{i,j}^{\mbox{\tiny{obj}}}\bigg(P_{i}^{j}\text{log}(\hat{P}_{i}^{j})+(1-P_{i}^{j})\text{log}(1-\hat{P}_{i}^{j})\bigg)\\
    \quad\quad-\lambda_{\mbox{\tiny{empty}}}\sum_{i=0}^{s\times s}\sum_{j=0}^{N}I_{i,j}^{\mbox{\tiny{empty}}}\bigg(P_{i}^{j}\text{log}(\hat{P}_{i}^{j})+(1-P_{i}^{j})\text{log}(1-\hat{P}_{i}^{j})\bigg)
\end{gather}

The total loss can be calculated
\begin{align}
    \mathcal{L}_{\mbox{\tiny{total}}} = \mathcal{L}_{\mbox{\tiny{box}}} + \mathcal{L}_{\mbox{\tiny{score}}} + \mathcal{L}_{\mbox{\tiny{cls}}}
\end{align}
and we seek parameters that minimise the total loss in the training phase.

\begin{figure}
\centering
\includegraphics[width=\textwidth]{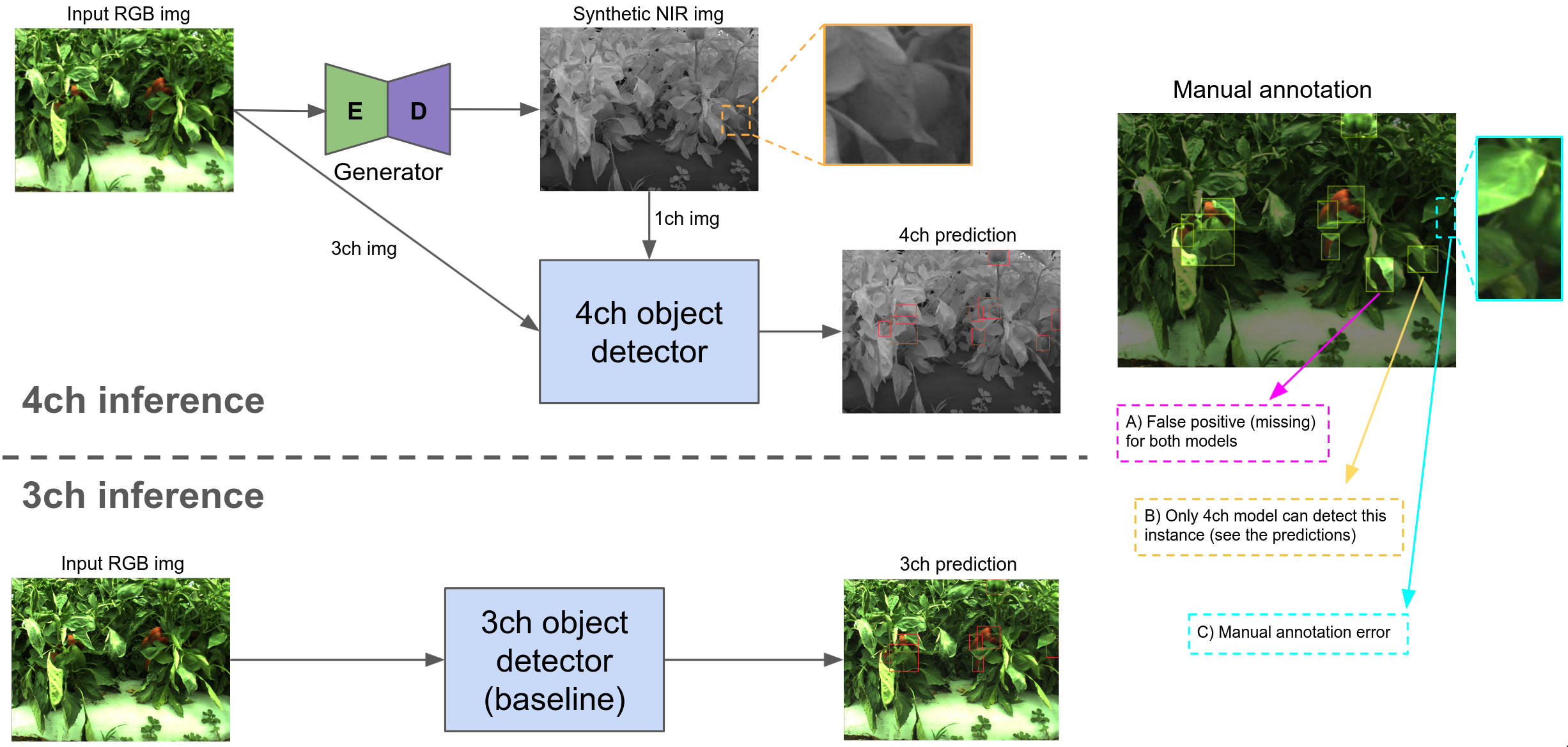}
\caption{4 channel early fusion object detection inference pipeline (top) and 3 channel ordinary inference (bottom). Brighter areas in manual annotation on the right image indicate manual labeled bounding boxes. One can compare them with model predictions (i.e., red boxes in '4ch prediction' and '3ch prediction'). This image best viewed in zoom-in.)}
\label{fig:object-detection}
\end{figure}

Figure \ref{fig:object-detection} illustrates the object detection pipeline we proposed in this paper. Among various visual fusion approaches, we follow `early fusion' in order to maintain similar inference processing time as of 3 channel inference (`late fusion' requires $O(N)$ complexity where $N$ is the number of input). Moreover, it is easier to implement and straight forward to extend from 3 channel baseline.
Firstly, input image $\in\mathbb{R}^{\text{WxHx3}}$ is fed into the generator that learnt mapping from visible-to-infrared domain and outputs a synthetic image $\in\mathbb{R}^{\text{WxHx1}}$. These two data are concatenated to form the shape of input data $\in\mathbb{R}^{\text{WxHx4}}$ prior to the input convolution layer (i.e., early fusion). After forward computation, the network predicts bounding boxes with the corresponding confidence (see red boxes in the figure). `4ch inference' and `3ch inference' indicate this prediction with synthetic image and without it, respectively. From this cherry-picked experiment, we observed interesting aspects 1) there is an instance that only the 4 channel model can detect. Marked in yellow from manual annotation, the 3 channel model missed (False Negative) a capsicum obscured by leaves and severe shadow. In comparison, the 4 channel model correctly detected it, and we believe this is the impact of introducing synthetic images. 2) Both models failed in very challenging instances (marked in magenta). 3) Both models successfully detect the object despite manual annotation error (marked in cyan).

\subsubsection{Datasets used for fruit detection (4ch)}

In order to see the impact of synthetically generated images, we created a dataset that includes 4 channel \textbf{11} fruits built upon our previous study \cite{Sa2016-zv} which presented 7 fruits detection. Blueberry, cherry, kiwi, and wheat are newly introduced in this dataset. Even though the total number of images is far less than other publicly available datasets such as ImageNet, COCO (except wheat), this may be useful for mode pre-training for another downstream task. Each image contains multiple instances because fruits usually form a cluster. In addition to this, each fruit image was taken in various camera views, scale, and lighting conditions which are very helpful for model generalisation. We made this dataset publicly available in a cloud annotation framework so that one can download them in many different formats\footnote{\url{https://github.com/inkyusa/deepNIR_dataset}}. Note that we manually generated and fixed errors in our previous dataset\cite{Sa2016-zv} except wheat dataset obtained from a machine learning competition\footnote{\url{https://www.kaggle.com/c/global-wheat-detection}}. Dataset split followed the 8:1:1 rule for train/validation/test and final object detection results were reported using the test set. Detailed experiments results and dataset samples are presented in the following experiments section.

\begin{table}
\caption{Dataset for object detection summary table\label{tbl:object-detection-dataset}}
\begin{center}
\begin{tabular}{ccccccc}
\hline
\multirow{2}{*}{\textbf{Name}} & \multicolumn{3}{c}{\textbf{\# images}}    &                & \multirow{2}{*}{\textbf{\# instances}} & \textbf{Median Image ratio} \\
                               & Train (80\%) & Valid (10\%) & Test (10\%) & \textbf{Total} &                                        & \textbf{(wxh)}     \\ \hline
apple                          & 49           & 7            & 7           & 63             & 354                                    & 852x666            \\
avocado                        & 67           & 7            & 10           & 84             & 508                                    & 500x500            \\
blueberry                      & 63           & 8            & 7           & 78             & 3176                                   & 650x600            \\
capsicum                       & 98           & 12           & 12          & 122            & 724                                    & 1290x960           \\
cherry                         & 123          & 15           & 16          & 154            & 4137                                   & 750x600            \\
kiwi                           & 100          & 12           & 13          & 125            & 3716                                   & 710x506            \\
mango                          & 136          & 17           & 17          & 170            & 1145                                   & 500x375            \\
orange                         & 52           & 6            & 8           & 66             & 359                                    & 500x459            \\
rockmelon                      & 77           & 9            & 11           & 97             & 395                                    & 1290x960           \\
strawberry                     & 63           & 7            & 9           & 79             & 882                                    & 800x655            \\
wheat                          & 2699         & 337          & 337         & 3373           & 147793                                 & 1024x1024         
\end{tabular}
\end{center}
\end{table}

\section{Experiments and results}\label{sec:results}
In this section, we first define evaluation metrics used for synthetically generated images and object detection tasks. Based on these, quantitative and qualitative results are presented for both synthetic NIR image generation and object detection tasks.

\subsection{Evaluation metrics}\label{subsec:eval-metric}
For synthetic image generation, it is relatively difficult to accurately gauge performance because the generator model often produces imaginary images (e.g., fake faces). Fortunately, in our task, we can utilise either traditional image similarity metrics or feature space image distribution comparisons because our objective is to generate a synthetic NIR image with a small residual error compared to the original NIR. For the object detection task, we adopt mean average precision with IoU sweeping range from [0.5:0.95] ($\text{mAP}_{\mbox{\tiny{[0.5:0.95]}}}$) with 0.05 step.

\subsubsection{Synthetic image evaluation metrics}
As mentioned with inherent challenges in evaluating synthetic images, communities widely have used various performance metrics\cite{Borji2018-ow} such as Frechet Inception Distance (FID) or  
FID, Generative Adversarial Metric (GAM). Each of them has its particular advantages and disadvantages. Among them, we choose FID which reports image similarity between two images in high-dimensional feature space. It implies the metric finds the distance between two multivariate Gaussian distributions, $X_A\sim\mathcal{N}(\mu_A, \Sigma_A)$ and $X_B\sim\mathcal{N}(\mu_B, \Sigma_B)$ that are fitted to data embedded into a feature space (e.g., extracted features using InceptionNet or VGG16 backbone).

\begin{align}
    d_{\mbox{\tiny{FID}}}(A,B) = ||\mu_A - \mu_B||_{2}^2+Tr(\Sigma_A+\Sigma_B-2\sqrt{\mu_A\mu_B})
\end{align}

\subsubsection{Object detection evaluation metric}
There are also many metrics for object detection tasks such as IoU, GIoU, mAP, and F-$\alpha$\cite{Padilla2020-fm}. Among them, $\text{mAP}_{\mbox{\tiny{[0.5:0.95]}}}$ is a widely utilised and acceptable metric\cite{Lin2014-eo} and it is defined as follow

\begin{align}
P =& \frac{TP}{TP+FP} \\
R =& \frac{TP}{TP+FN} \\
AP =& \int_0^1P(R)\delta R \\
mAP =& \frac{1}{M}\sum_{i=1}^{M}\text{AP}_i
\end{align}
where $TP$, $FP$, $FN$ denote true-positive, false-positive, and false-negative respectively. True positive implies our prediction is correct as of annotation bounding box (hit), false positive is when we wrongly make a prediction (false-alarm), and false-negative occurs when we miss a bounding box (miss). Note that true negative $TN$ is not considered in object detection task because this implies correct rejection (e.g., there should not be an bounding box and a model does not predict at that location) and there exist infinite cases that satisfy the condition. $P$ and $R$ are precision and recall, and AP and mAP refers to average precision, and mean average precision that is the mean of all class's AP. In our case, AP and mAP are treated equally because we only have one class ($M=1$) in our train dataset. As shown, AP is equal to the area of precision-recall curve and $\text{mAP}_{\mbox{\tiny{[0.5:0.95]}}}$ is mean average precision of all classes while varying IoU threshold range from 0.5 to 0.95 with 0.05 steps (i.e., mean of total 20 samples).

In terms of train/inference processing time per dataset, and GPU devices used for each task, Table~\ref{tbl:object-detection-dataset}
\begin{table}
\caption{Train/inference summray table\label{tbl:train-inference-time}}
\label{tbl:object-detection-dataset}
\begin{tabular}{ccccccccc}
\textbf{Type}                                                                             & \textbf{Name}              & \textbf{Desc.} & \textbf{GPU}                                                        & \textbf{epoch} & \textbf{\begin{tabular}[c]{@{}c@{}}Train \\ (hrs)\end{tabular}}                  & \textbf{\begin{tabular}[c]{@{}c@{}}Test \\ (ms/image)\end{tabular}} & \textbf{\begin{tabular}[c]{@{}c@{}}Img \\ size\end{tabular}}           & \textbf{\# samples}              \\ \hline
\multirow{7}{*}{\begin{tabular}[c]{@{}c@{}}Synthetic \\ image \\ generation\end{tabular}} & \multirow{4}{*}{nirscene}  & 10x            & \multirow{4}{*}{P100}                                               & 300            & 8                                                                                & \multirow{4}{*}{500}                                                & \multirow{4}{*}{\begin{tabular}[c]{@{}c@{}}w:256\\ h:256\end{tabular}} & 2880                             \\
                                                                                          &                            & 100x           &                                                                     & 300            & 72                                                                               &                                                                     &                                                                        & 14400                            \\
                                                                                          &                            & 200x           &                                                                     & 300            & 144                                                                              &                                                                     &                                                                        & 28800                            \\
                                                                                          &                            & 400x           &                                                                     & 164            & 113                                                                              &                                                                     &                                                                        & 57600                            \\ \cline{2-9} 
                                                                                          & \multirow{2}{*}{SEN12MS}   & All            & \multirow{2}{*}{\begin{tabular}[c]{@{}c@{}}RTX\\ 3090\end{tabular}} & 112            & 58                                                                               & \multirow{2}{*}{250}                                                & \multirow{2}{*}{\begin{tabular}[c]{@{}c@{}}w:256\\ h:256\end{tabular}} & 144528                           \\
                                                                                          &                            & Summer         &                                                                     & 300            & 140                                                                              &                                                                     &                                                                        & 36601                            \\ \cline{2-9} 
                                                                                          & \multicolumn{2}{c}{Capsicum}                & \begin{tabular}[c]{@{}c@{}}RTX\\ 3090\end{tabular}                  & 300            & 29                                                                               & 780                                                                 & \begin{tabular}[c]{@{}c@{}}w:1280\\ h:960\end{tabular}                 & 129                              \\ \hline
\multirow{4}{*}{\begin{tabular}[c]{@{}c@{}}Fruit\\ detection\end{tabular}}                & \multirow{2}{*}{10 Fruits} & Yolov5s        & \multirow{2}{*}{\begin{tabular}[c]{@{}c@{}}RTX\\ 3090\end{tabular}} & 600            & \multirow{2}{*}{\begin{tabular}[c]{@{}c@{}}min: 4min,\\ max: 36min\end{tabular}} & 4.2-10.3                                                            & \begin{tabular}[c]{@{}c@{}}w:640\\ h:640\end{tabular}                  & \multirow{2}{*}{\textless{}5000} \\
                                                                                          &                            & Yolov5x        &                                                                     &                &                                                                                  &                                                                     &                                                                        &                                  \\ \cline{2-9} 
                                                                                          & \multirow{2}{*}{Wheat}     & Yolov5s        & \multirow{2}{*}{\begin{tabular}[c]{@{}c@{}}RTX\\ 3090\end{tabular}} &                & 1h 16min                                                                         & 4.2-10.3                                                            & \begin{tabular}[c]{@{}c@{}}w:640\\ h:640\end{tabular}                  & \multirow{2}{*}{147793}          \\
                                                                                          &                            & Yolov5x        &                                                                     &                & 2h 16min                                                                         &                                                                     &                                                                        &                                 
\end{tabular}
\end{table}

\subsection{Quantitative results for synthetic NIR image generation}
In this section, we present three quantitative synthetic image generation results for nirscene1, SEN12MS and capsicum datasets.

An L. \textit{et al.} demonstrated impressive results by making use of the multi-channel attention selection module\cite{An2019-xl}. They cropped 3691 images with 256$\times$256 resolution for model training and testing. Unfortunately, the dataset used in this study is unavailable and technical details are insufficient for fair comparisons (e.g., train/test samples and their split is not disclosed). 

As a rule of thumb, we split our dataset 8:1:1 for train/validation/test as described in Table \ref{tbl:synthetic-dataset} and proceeded with experiments. We achieved comparable results as shown in Table \ref{tbl:nirscene1-quantitative-results}. It can be observed that FID keeps improving (the lower is the better), corresponding to oversampling rate. It is a fact that the maximum amount of information from the oversampled samples should be less than or equal to the original dataset. This redundancy may introduce system overhead as mentioned in \cite{Birodkar2019-nv} and make a marginal impact on image segmentation or object detection tasks. However, training the GAN-style model often demonstrates the distinct characteristics as follows; speaking of convergence, it is often difficult to find the optimal point for GAN models because of its naturally inherent min-max game framework. Even with stable convergence, it is commonly difficult to guarantee the performance of the trained GAN model due to mode-collapse or diminished gradient issues \cite{Kodali2017-jf}. If the number of samples is small (e.g., $<$10\unit{k}), it is very challenging to train a stable model.

By oversampling, we hypothesise that GAN models can learn stable parameters significantly affected by the number of samples, especially batch-normalisation layers. This operation, namely hard-cropping, can also be done during the training phase by utilising augmentation approaches (soft-cropping). However, this soft cropping often is performed based on user-defined probability and may introduce more instability. The hard-cropping effect can be achieved with the maximum probability and a large number of epochs. 

\begin{table}
\caption{Synthetic image generation quantitative results}\label{tbl:nirscene1-quantitative-results}
\begin{center}
\begin{tabular}{ccccccc}
                      & \textbf{FID}$\downarrow$ & \textbf{\# Train}& \textbf{\# Valid} & \textbf{\# Test}& \textbf{img size}&\textbf{Desc.} \\ \hline
\cite{An2019-xl}               & 28     &3691&N/A&N/A &256$\times$256     & \begin{tabular}[c]{@{}c@{}}10 layer attentions,\\ 16 pixels, 4 layers,\\ encoding+decoding\end{tabular} \\ \cline{1-7} 
\multirow{5}{*}{nirscene1} & 109.25     &2880&320&320&256$\times$256  & x10 oversample                                                                                          \\
                      & 42.10   &14400&1600&1700 &256$\times$256    & x100 oversample                                                                                         \\
                      & 32.10   &28800&3200&3400 &256$\times$256    & x200 oversample                                                                                         \\
                      & 27.660  &57600&6400&6800 &256$\times$256    & x400 oversample, epoch 89
                                                                                         \\ 
                      & \textbf{26.53}   &86400&9600&9600&256$\times$256    & x400 oversample, epoch 114
                      \\ \hline
\multirow{2}{*}{SEN12MS}& 16.47& 36601&4576&4576&256$\times$256&Summer, 153 epoch \\
                        & \textbf{11.36}& 144528&18067&18067&256$\times$256&All season, 193 epoch \\ \cline{1-7} 
       
capsicum& 40.15& 1102&162&162& 1280$\times$960&150 epoch \\
\end{tabular}
\end{center}
\end{table}

To our best knowledge, An L. \textit{et al.}\cite{An2019-xl} is the only comparable baseline that exploited the nirscene1 dataset reporting FID score. However, this study inaccurately describes essential technical details such as how many test images were evaluated and data split. Therefore, it is difficult to make a fair comparison between our results and \cite{An2019-xl}. 

Yuan X. \textit{et al.}\cite{Yuan2020-ma} reported impressive results using the SEN12MS dataset for synthetic image generation. 30000/300 images were randomly sampled from the Summer dataset for training and testing, respectively. They reported quantitative results with image similarity metrics such as mean absolute error (MAE) or structural similarity (SSIM), which differ from our evaluation metrics. More importantly, the split strategy is questionable as only 1\% was evaluated. The other 99\% of the dataset was utilised for the training. From our perspective, with such small test samples it is difficult to evaluate the model performance properly.

The capsicum data reported 40.15 FID score. This dataset is the closest range among all datasets and contains cluttered structures and complex scenes. It is difficult for models to learn RGB to NIR mapping (or vice-versa) properly.

We summarise from these quantitative results that the number of samples is significant for the GAN model.. Our synthetic NIR generator worked best for SEN12MS which performed proper radiometric calibration to have access to reflectances rather than raw pixel value. Accessing these reflectances is critical because they can hold consistent values despite acceptable changes in camera intrinsic or extrinsic parameters if images were taken under a similar light source. It is worth mentioning that image resolution is also one of the interesting aspects to consider, as shown in the higher-resolution result. If an image has a higher resolution, the more difficult for a network to learn the NIR-RGB relationship. In order to improve this, we are required to design a deeper network with more training samples.

The final analysis point is that the FID score is unitless and does not have a quality measure. A low score means high performance, but it is questionable whether 15 FID is good or bad. We suggest conducting a visual inspection of model prediction to address this issue, presented in the following section.

\subsection{Qualitative results for synthetic NIR image generation}
\begin{figure}[H]
\centering
\includegraphics[width=\textwidth]{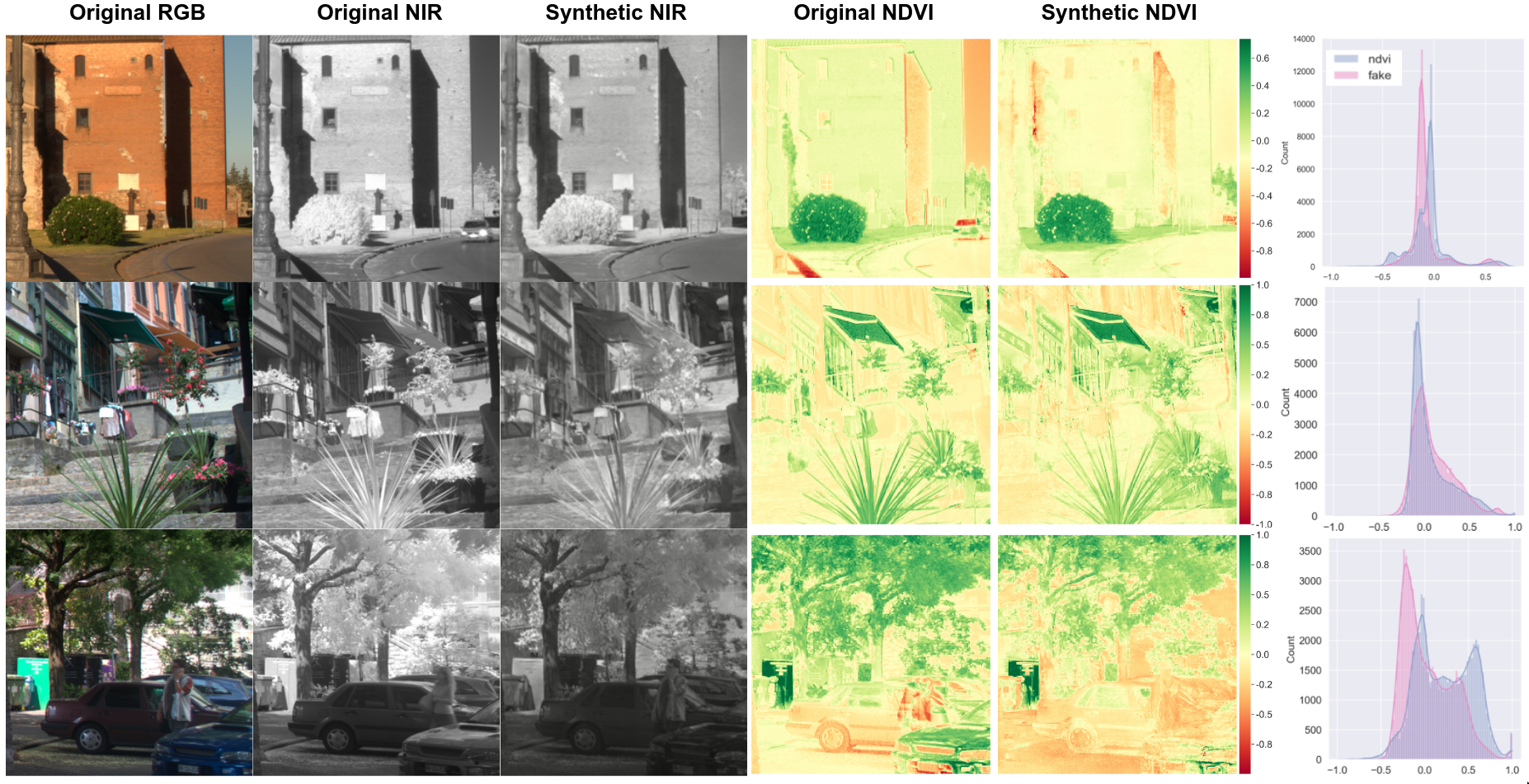}
\includegraphics[width=\textwidth]{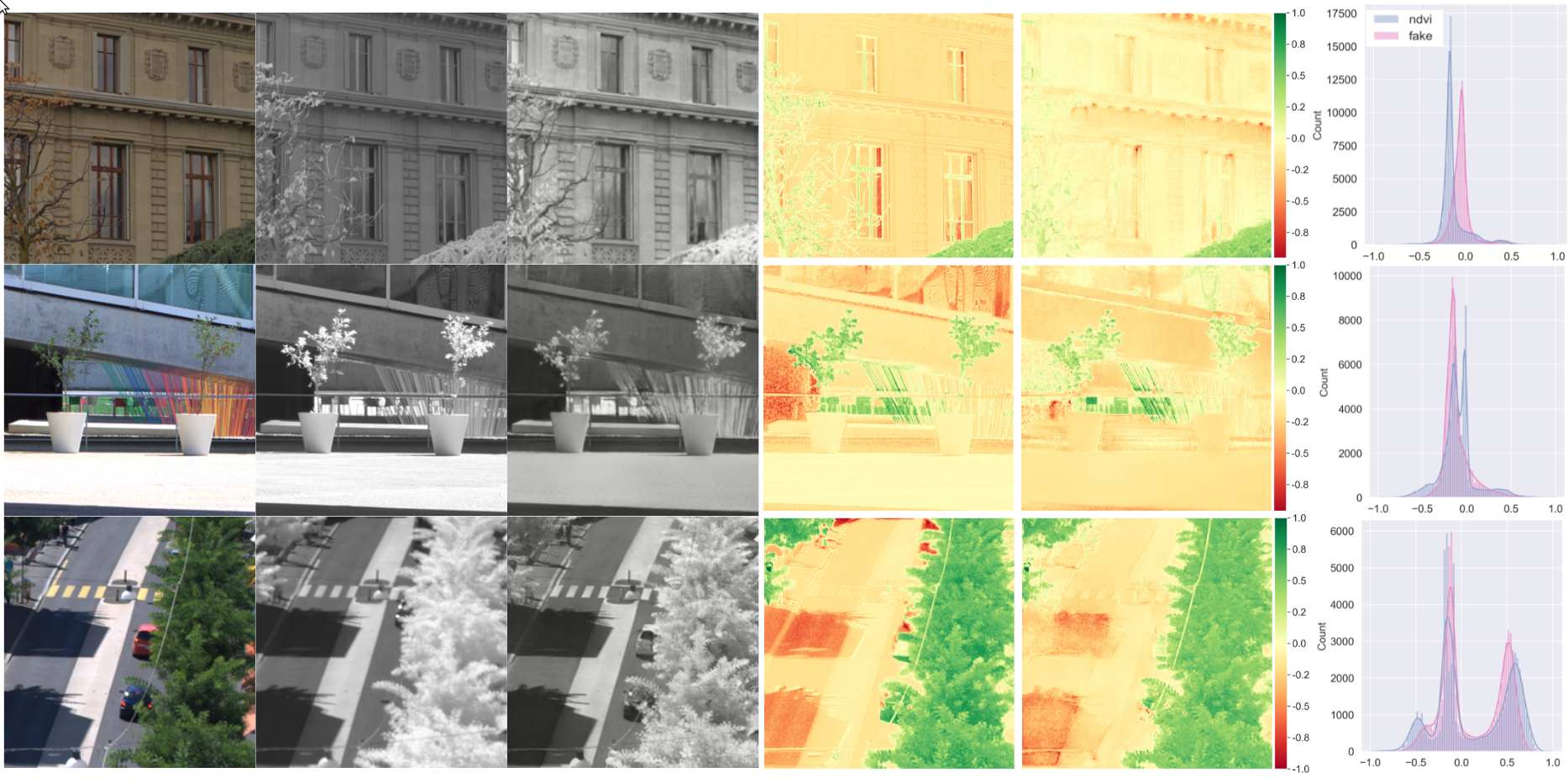}
\caption{6 samples of nirscene1 dataset (1st and 2nd columns) and their corresponding synthetic NIR image (3rd), and NDVI images generated from original and snythetic NIR images (4th and 5th). The right most column shows pixel distribution for each NDVI images (red: snythetic, blue: original).}
\label{fig:nirscene1-qualitative}
\end{figure}

We show qualitative results of 3 datasets, nirscene1, SEN12MS, and capsicum. The nirscene1 dataset holds FID of 26.53 and Figure \ref{fig:nirscene1-qualitative} exemplifies 6 randomly selected test samples. The leftmost two columns are the original NIR-RGB pair, and Synthetic NIR refers to model prediction ($\widehat{\text{NIR}}$) that resulted in the mentioned FID score. Original NDVI indicates extracted NDVI $\frac{NIR - RED}{NIR + RED}$ using original NIR-RGB pair and synthetic NDVI is output by calculating $\frac{\widehat{\text{NIR}} - RED}{\widehat{\text{NIR}} + RED}$. These two NDVI images' histogram is presented at the rightmost column (blue=original NDVI, red=synthetic). Generally speaking, the network learnt good nonlinear mapping. However, it shows limitations especially varying intensities and blurred edges. These are clearly noted from the histogram. The nirscene1 dataset contains many images taken outdoors where the light condition is inconsistent. This poses challenging conditions for the model to learn the mapping given a small dataset.

\begin{figure}[H]
\centering
\includegraphics[width=\textwidth]{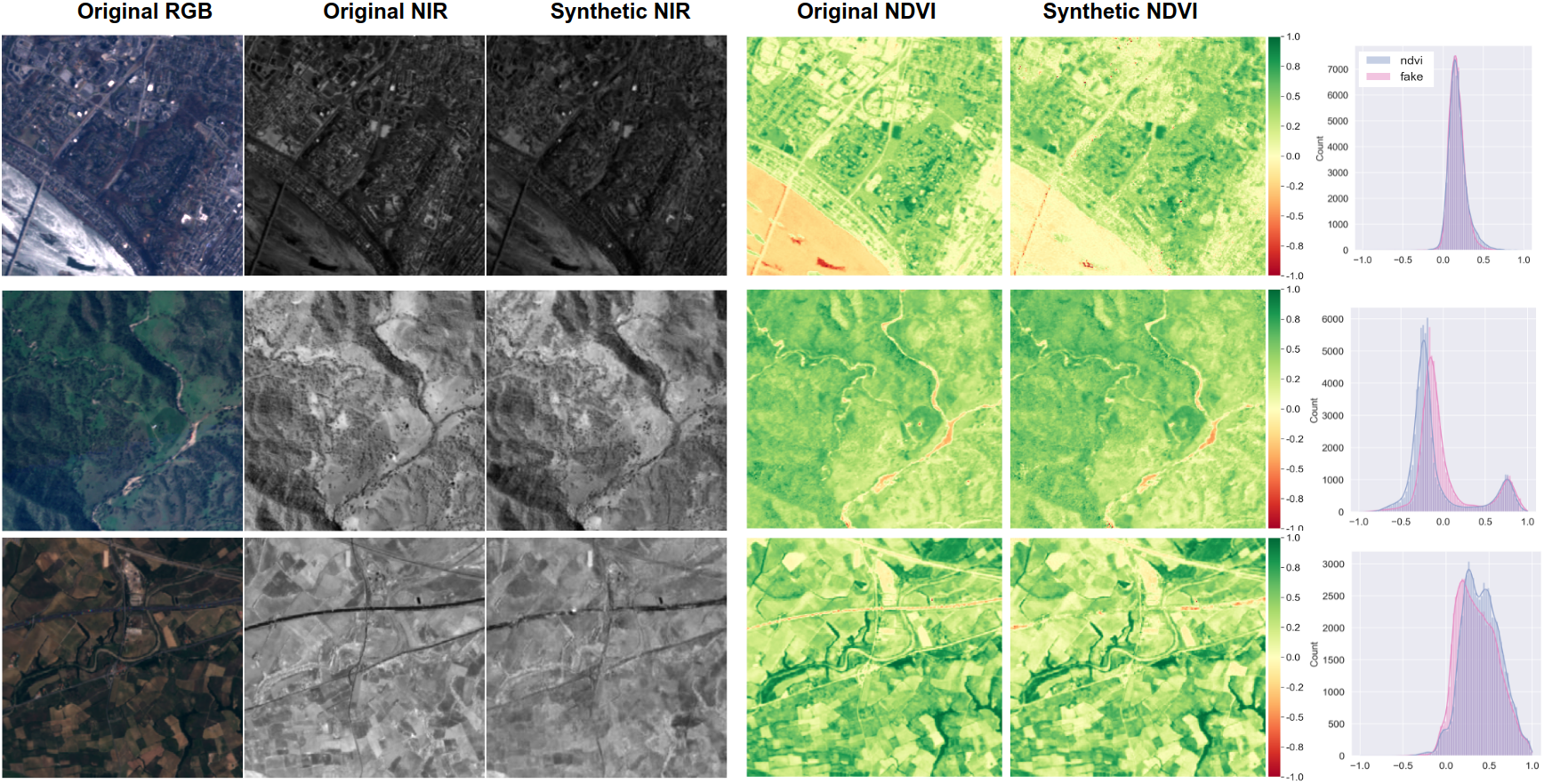}
\includegraphics[width=\textwidth]{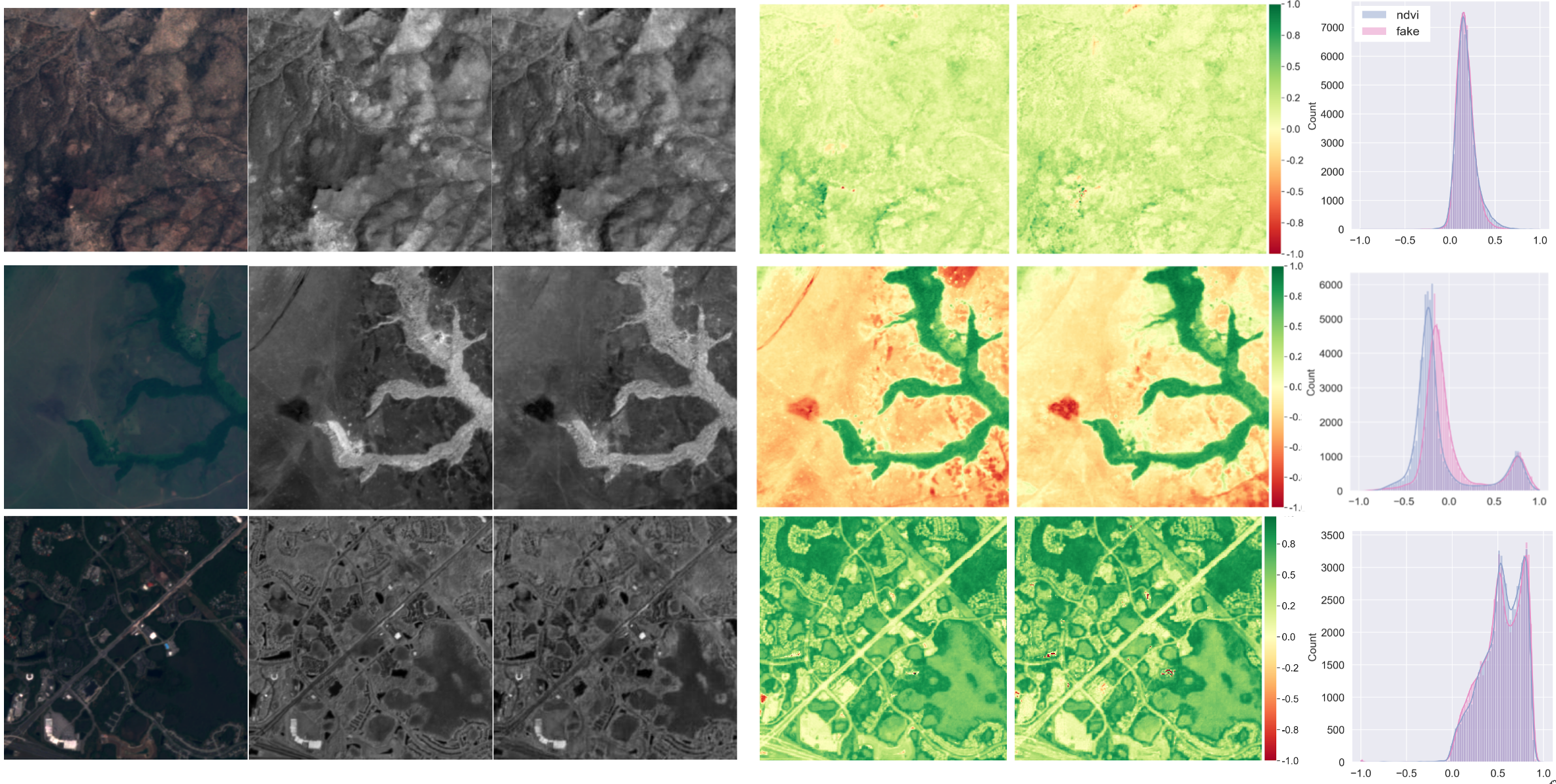}
\caption{6 samples of SEN12MS dataset (1st and 2nd columns) and their corresponding synthetic NIR image (3rd), and NDVI images generated from original and snythetic NIR images (4th and 5th). The right most column shows pixel distribution for each NDVI images (red: snythetic, blue: original).}
\label{fig:sentinel2-qualitative}
\end{figure}

SEN12MS dataset was taken/calibrated in more stable conditions than nirscene1. This clearly can be observed from Figure \ref{fig:sentinel2-qualitative}. Prediction nicely fits the original NDVI except for a couple of under-estimated points. We achieved an FID of 16.47 for the Summer and 11.36 for All season subsets in these experiments. Qualitatively speaking, it can be said that less than 15 FID is an excellent approximation of the original data. We are still investigating the reason for the underestimation. However, it is maybe a valid hypothesis that the network requires to see more images containing water as it mostly causes an error at very low NDVI, which is close to the normalised difference water index (NDWI) wavelength. 

\begin{figure}[H]
\centering
\includegraphics[width=\textwidth]{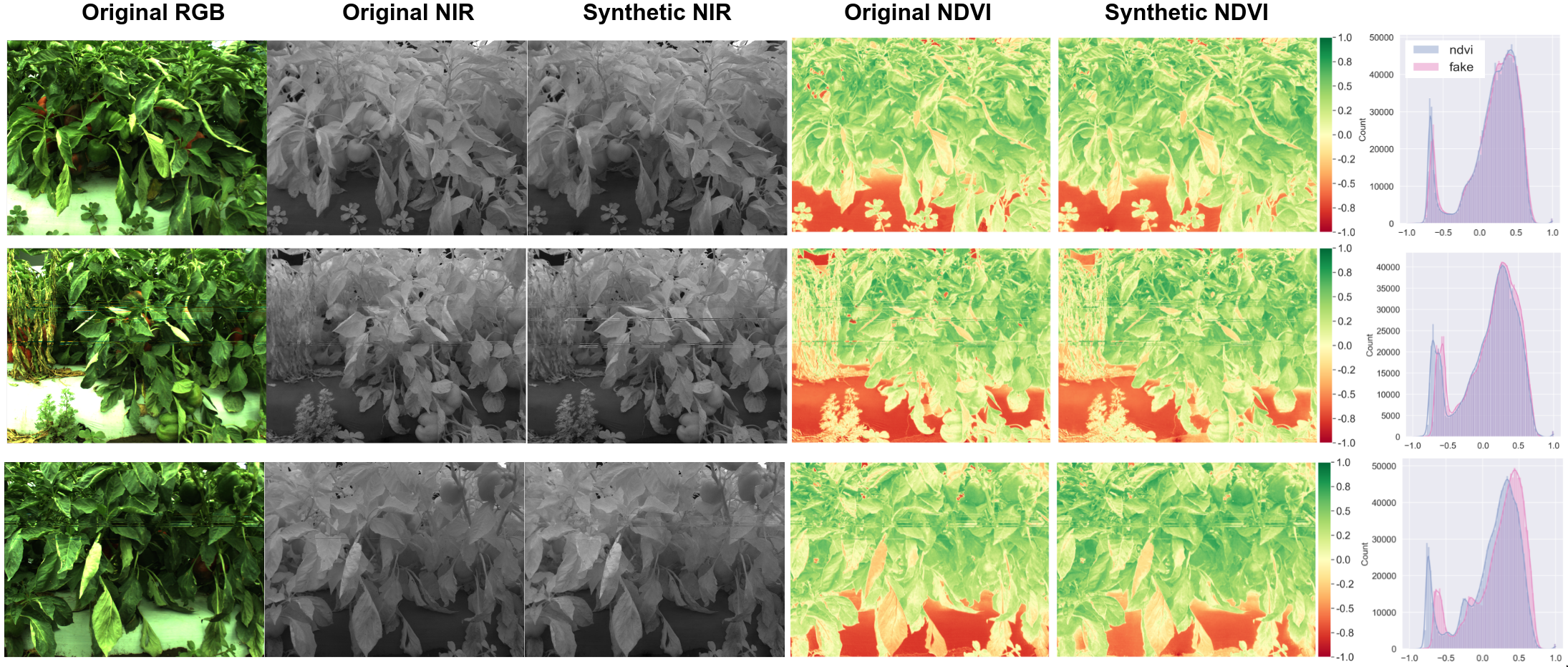}
\includegraphics[width=\textwidth]{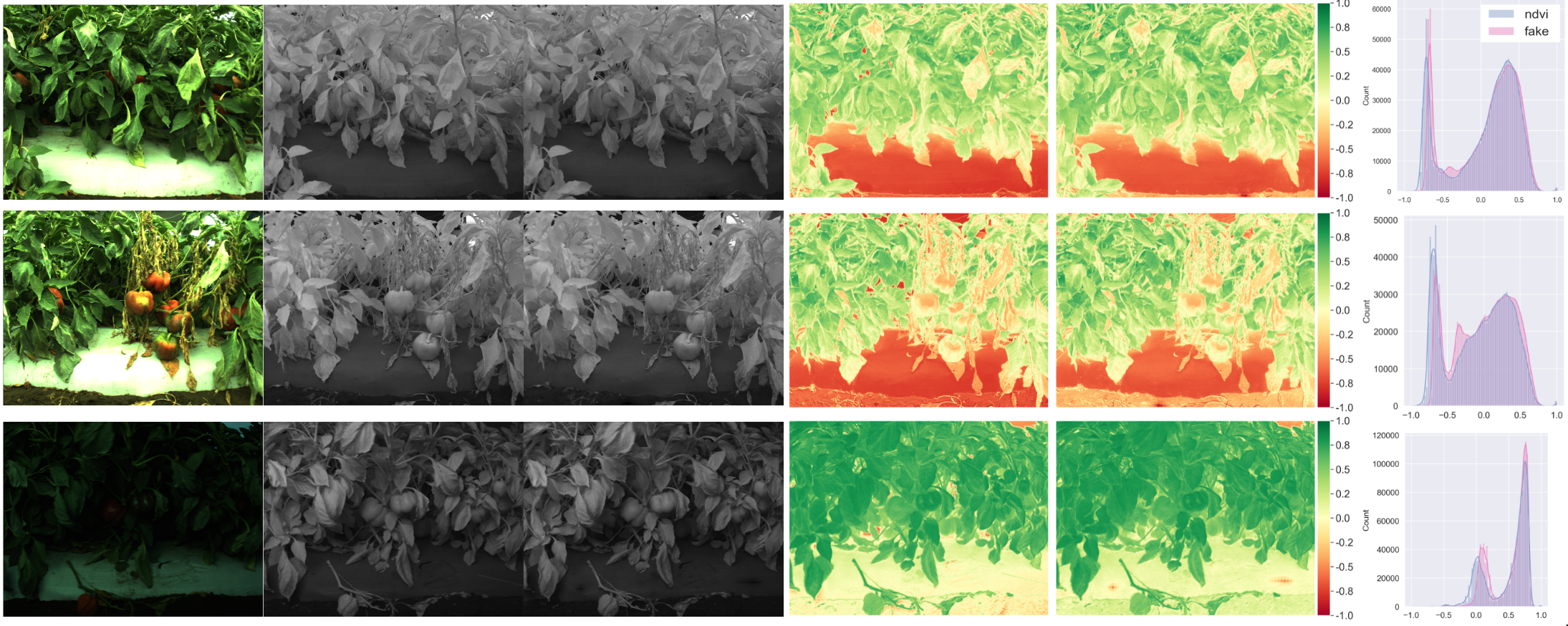}
\caption{6 samples of capsicum dataset (1st and 2nd columns) and their corresponding synthetic NIR image (3rd), and NDVI images generated from original and snythetic NIR images (4th and 5th). The right most column shows pixel distribution for each NDVI images (red: snythetic, blue: original).}
\label{fig:capsicum-qualitative}
\end{figure}

The last dataset is capsicum, as shown in Figure \ref{fig:capsicum-qualitative}. This dataset holds consistent illumination and white balance, but some samples are severely under-exposed (see the bottom row in the figure). In addition, this dataset was taken in the closest range, causing very cluttered and complex scenes. Although our model worked surprisingly well, it still could not recover sharp detail that led to a relatively higher FID of 40.15 but still impressive, and we can use this result to see if we can improve the object detection task.

\subsection{Applications for fruit detection using synthetic NIR and RGB images}
Synthetic NIR images (700-800\unit{nm} wavelength) can provide useful information that visible-range RGB images can not cover. One of the prominent properties is high reflectance in this particular bandwidth from vegetation due to chlorophyll in cells of leaves. Combining NIR with RED channel (i.e., Normalised Difference Vegetation Index (NDVI)I\cite{Rouse1973-oz}) enables the measurement of vegetated areas and their condition easily. Therefore, additional NIR information can significantly improve the performance of plant segmentation tasks as demonstrated in our previous studies \cite{Sa2018-fu}\cite{Sa2018-sg}.

Not only for the image segmentation task, but it can boost object detection performance as depicted in our previous experiments \cite{Sa2016-zv}. The provided information helps to enhance distinguishing power by providing quality features. For example, texture and objects under shallow shadow appear clearer in the infrared range than visible.

Thus, in this section, we aim at improving object detection performance by injecting additional synthetic data that we generated in the previous section. All experiments are conducted using the dataset mentioned in Table \ref{tbl:object-detection-dataset}. We adopt Yolov5 single stage detector for the experiments and made minor modifications in order to take 4 channel input. Any other object detectors such as Detectron2, MMDetection, or other Yolo-series can be easily utilised.

\subsection{Quantitative fruit detection results}
As mentioned in evaluation metric section (\ref{subsec:eval-metric}), $\text{mAP}_{\mbox{\tiny{[0.5:0.95]}}}$ is the key performance metric in this quantitative evaluation. 11 fruits and crops are considered as shown in Table \ref{tbl:object-detection-results}. Note that 7 fruits such as [apple, avocado, capsicum, mango, orange rockmelon, strawberry] were adopted from our previous work\cite{Sa2016-zv} when a two-stage detector (Faster-RCNN\cite{Ren2015-no}) was utilised as the main object detector. Although the samples for these 7 fruits remain almost identical, we re-annotated wrongly annotated samples and made all of them available via cloud service with the newly added 4 fruits dataset. This allows users to export the dataset in various formats suitable for seamlessly using various object detection frameworks with minimal effort.

We considered two models for each fruit; yolov5s (7.2M parameters) and yolov5x (86.7M), and each model is trained with/without synthetic NIR images. Therefore each fruit holds 4 performance results. All hyperparameters used are default from the latest repository, only the number of the epoch is set as 600 except for wheat, mainly due to being an order of magnitude larger in dataset size. Bolder indicates each fruit's best score corresponding to the metric. During the training phase, we assume only one class exists in a dataset which is a valid assumption considering mono-species fruit farm scenarios. In generating synthetic NIR images, we deployed the generator trained with the capsicum dataset mentioned in Table \ref{tbl:synthetic-dataset}. It is worth mentioning that two capsicum datasets used for synthetic NIR generation and object detection are different in various aspects such as data collection campaign location, time, and lighting condition. This is because we can only obtain capsicum annotations from our previous work\cite{Sa2016-zv} for object detection, whereas we have abundant un-annotated RGB+NIR pairs collected from trial collections for synthetic NIR generation.   

Overall all detection performance is impressive despite a small number of train samples. $\text{mAP}_{\mbox{\tiny{0.5}}}$ are reported min-max of [0.85-0.98] and $\text{mAP}_{\mbox{\tiny{[0.5:0.95]}}}$ of [0.49-0.81]. 4 fruits such as apple, capsicum, avocado, orange outperform by making use of additional NIR information, 7 others report the best performance only with RGB information. This result points against our objective, and we would like to elaborate on causality deeply..

Stable and consistent reflectance plays a significantly important role in synthetic image generation. Intuitively, this implies our network is required to learn an RGB to NIR nonlinear mapping with small variations. If distributions and characteristics in the dataset significantly vary, our model would require more datasets covering the envelope. Otherwise, it will be under-, over-fitted that in turn leads to inferior performance that occurred in our case. All datasets evaluated in object detection have a marginal correlation with the capsicum dataset utilised for GAN model learning. Many of them were obtained from web pages without NIR images. More detailed limitations, failure cases and possible workarounds are discussed in the next section \ref{sec:discussion}.

\begin{table}
\begin{center}
\caption{11 fruits/crops object detection quantitative results table. Up-arrow indicates a higher score is better performance. Bold denotes the best performance in the corresponding metric within each fruit.}\label{tbl:object-detection-results}
\begin{tabular}{cccccccc}
\hline \hline
\textbf{Name}               & \textbf{Model}           & \textbf{Input} & \textbf{$\text{mAP}_{\mbox{\tiny{0.5}}}$$\uparrow$} & \textbf{$\text{mAP}_{\mbox{\tiny{[0.5:0.95]}}}$}$\uparrow$ & \textbf{precision}$\uparrow$ & \textbf{recall}$\uparrow$ & \textbf{F1}$\uparrow$\\ \hline
\multirow{4}{*}{apple}      & \multirow{2}{*}{yolov5s} & RGB            & 0.9584           & 0.7725                &0.9989 &0.913&0.9540\\
                            &                          & RGB+NIR        & \textbf{0.9742}           & \textbf{0.7772} &0.9167 &0.9565&0.9362                \\ \cline{2-8}
                            & \multirow{2}{*}{yolov5x} & RGB            & 0.9688           & 0.7734  &0.9989 &0.9130&0.9540             \\
                            &                          & RGB+NIR        & 0.9702           & 0.7167    &0.8518 &0.9999&0.9199            \\ \hline
\multirow{4}{*}{avocado}    & \multirow{2}{*}{yolov5s} & RGB            & 0.8419           & 0.4873 &0.9545 &0.8077&0.8750               \\
                            &                          & RGB+NIR        & 0.8627           & 0.4975      &0.8749 &0.8071&0.8396            \\ \cline{2-8}
                            & \multirow{2}{*}{yolov5x} & RGB            & \textbf{0.9109}           & 0.6925   &1.0000 &0.8461&0.9166               \\
                            &                          & RGB+NIR        & 0.8981           & \textbf{0.6957}   &0.9583 &0.8846&0.9200               \\ \hline
\multirow{4}{*}{blueberry}  & \multirow{2}{*}{yolov5s} & RGB            & \textbf{0.9179}           & \textbf{0.5352}  &0.8941 &0.9018&0.8979                \\
                            &                          & RGB+NIR        & 0.8998           & 0.5319     &0.9224 &0.8354&0.8767             \\ \cline{2-8}
                            & \multirow{2}{*}{yolov5x} & RGB            & 0.9093           & 0.5039    &0.9345 &0.8494&0.8899              \\
                            &                          & RGB+NIR        & 0.8971           & 0.4657  &0.9063 &0.8476&0.8760                \\ \hline
\multirow{4}{*}{capsicum}   & \multirow{2}{*}{yolov5s} & RGB            & 0.8503           & 0.4735  &0.8159 &0.8473&0.8313                \\
                            &                          & RGB+NIR        & 0.8218           & 0.4485  &0.848 &0.8091&0.8281                \\ \cline{2-8}
                            & \multirow{2}{*}{yolov5x} & RGB            & 0.8532           & \textbf{0.4909}  &0.8666 &0.8429&0.8546                \\
                            &                          & RGB+NIR        & \textbf{0.8642}           & 0.4812   &0.8926 &0.8244&0.8571               \\ \hline
\multirow{4}{*}{cherry}     & \multirow{2}{*}{yolov5s} & RGB            & 0.9305           & 0.6045  &0.9300 &0.8586&0.8929                \\
                            &                          & RGB+NIR        & 0.9034           & 0.5655    &0.8994 &0.8505&0.8743              \\ \cline{2-8}
                            & \multirow{2}{*}{yolov5x} & RGB            & \textbf{0.9415}           & \textbf{0.6633}    &0.929	&0.8747	&0.9010              \\
                            &                          & RGB+NIR        & 0.9300           & 0.6325     &0.9129	&0.8687	&0.8903             \\ \hline
\multirow{4}{*}{kiwi}       & \multirow{2}{*}{yolov5s} & RGB            & 0.8642           & 0.5651    &0.9196	&0.7831	&0.8459              \\
                            &                          & RGB+NIR        & 0.8154           & 0.5195      &0.9039	&0.7188	&0.8008            \\ \cline{2-8}
                            & \multirow{2}{*}{yolov5x} & RGB            & \textbf{0.8935}           & \textbf{0.6010}     &0.9056	&0.8326	&0.8676              \\
                            &                          & RGB+NIR        & 0.8240           & 0.5139    &0.8625	&0.7643	&0.8104              \\ \hline
\multirow{4}{*}{mango}      & \multirow{2}{*}{yolov5s} & RGB            & 0.9431           & 0.6679  &0.9516	&0.879	&0.9139                \\
                            &                          & RGB+NIR        & 0.9032           & 0.6033    &0.9347	&0.8217	&0.8746              \\ \cline{2-8}
                            & \multirow{2}{*}{yolov5x} & RGB            & \textbf{0.9690}            & \textbf{0.6993}     &0.9333	&0.8917	&0.9120             \\
                            &                          & RGB+NIR        & 0.9339           & 0.6681     &0.9221	&0.9044	&0.9132             \\ \hline
\multirow{4}{*}{orange}     & \multirow{2}{*}{yolov5s} & RGB            & 0.9647           & 0.707     &0.9409	&0.9697	&0.9551             \\
                            &                          & RGB+NIR        & 0.9488           & 0.7669   &0.9655	&0.8482	&0.9031               \\ \cline{2-8}
                            & \multirow{2}{*}{yolov5x} & RGB            & \textbf{0.9662}           & \textbf{0.8484}  &0.9998	&0.9091	&0.9523                \\
                            &                          & RGB+NIR        & 0.9584           & 0.8277  &1.0000	&0.9091	&0.9524                \\ \hline
\multirow{4}{*}{rockmelon}  & \multirow{2}{*}{yolov5s} & RGB            & 0.9588           & 0.6321   &0.9198	&0.8846	&0.9019               \\
                            &                          & RGB+NIR        & 0.9205           & 0.6701      &0.9999	 &0.8462	&0.9167            \\ \cline{2-8}
                            & \multirow{2}{*}{yolov5x} & RGB            & \textbf{0.9612}           & \textbf{0.7161}    &0.9259	&0.9615	&0.9434              \\
                            &                          & RGB+NIR        & 0.9444           & 0.7018      &0.8926	&0.9615	&0.9258            \\ \hline
\multirow{4}{*}{strawberry} & \multirow{2}{*}{yolov5s} & RGB            & \textbf{0.9553}           & \textbf{0.6995}  &0.9559	&0.8784	&0.9155                \\
                            &                          & RGB+NIR        & 0.8913           & 0.6210     &0.9000	&0.8513	&0.8750             \\ \cline{2-8}
                            & \multirow{2}{*}{yolov5x} & RGB            & 0.8899           & 0.5237  &0.8954	&0.8108	&0.8510                \\
                            &                          & RGB+NIR        & 0.9071           & 0.4882    &0.9375	&0.8106	&0.8694              \\ \hline
\multirow{4}{*}{wheat}      & \multirow{2}{*}{yolov5s} & RGB            & 0.9467           & 0.5585     &0.929	&0.9054	&0.9170             \\
                            &                          & RGB+NIR        & 0.9412           & 0.5485    &0.9258	&0.8926	&0.9089              \\ \cline{2-8}
                            & \multirow{2}{*}{yolov5x} & RGB            & \textbf{0.9472}           & \textbf{0.5606}     &0.9275	&0.9035	&0.9153             \\
                            &                          & RGB+NIR        & 0.9294           & 0.5329  &0.9163	&0.9005	&0.9083  \\
\bottomrule
\end{tabular}
\end{center}
\end{table}

\begin{figure}
\centering
\includegraphics[width=\textwidth]{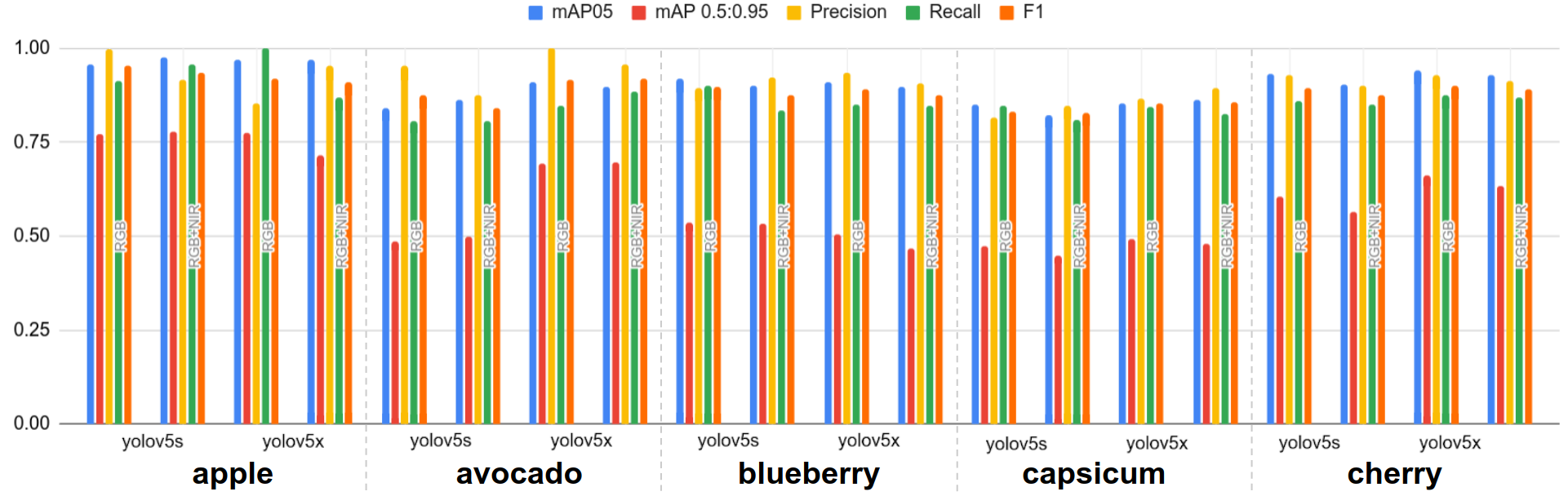}
\includegraphics[width=\textwidth]{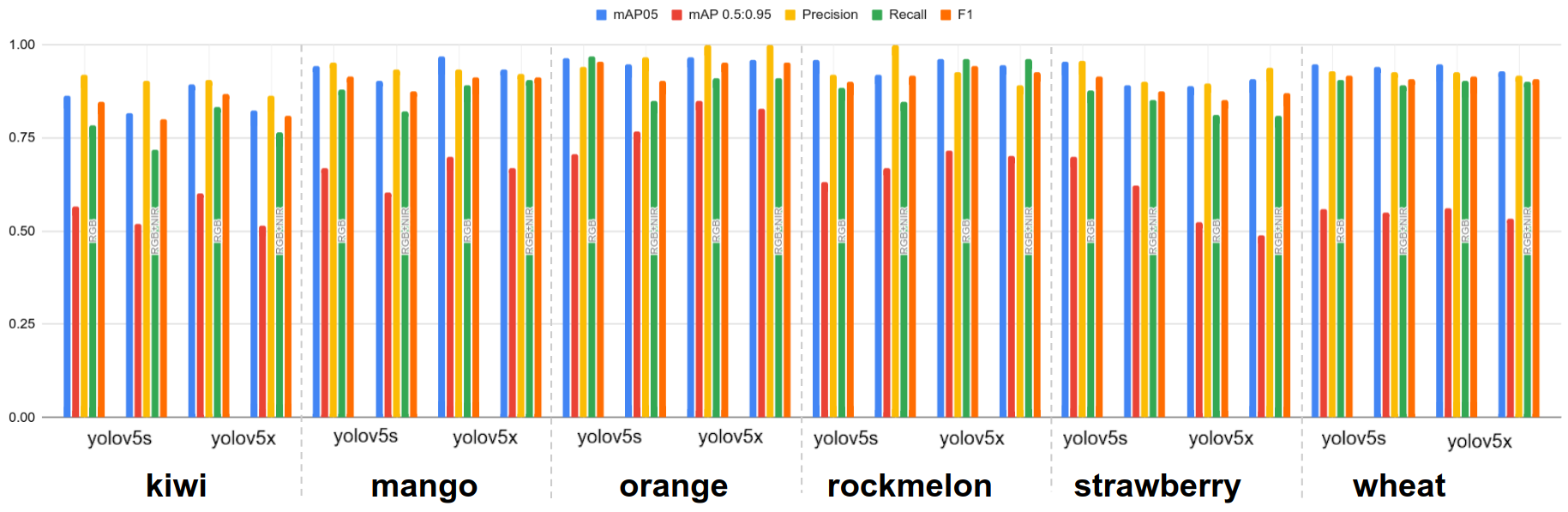}
\caption{Object detection results summary. Different colours indicate the corresponding metrics. The different type of input data (i.e., RGB or RGB+NIR) are separately grouped for each yolov5 models.}
\label{fig:detection results}
\end{figure}

Figure \ref{fig:detection results} shows a different view of Table \ref{tbl:object-detection-results}. It is clear that a model with more parameters performs better as cost of longer training time and hardware resources. At a first glance, the performance gap between only RGB and RGB+NIR is difficult to distinguish. To our best knowledge, it is sufficient to exploit only RGB images for fruits bounding box detection.

\begin{figure}
\centering
\includegraphics[width=0.8\textwidth]{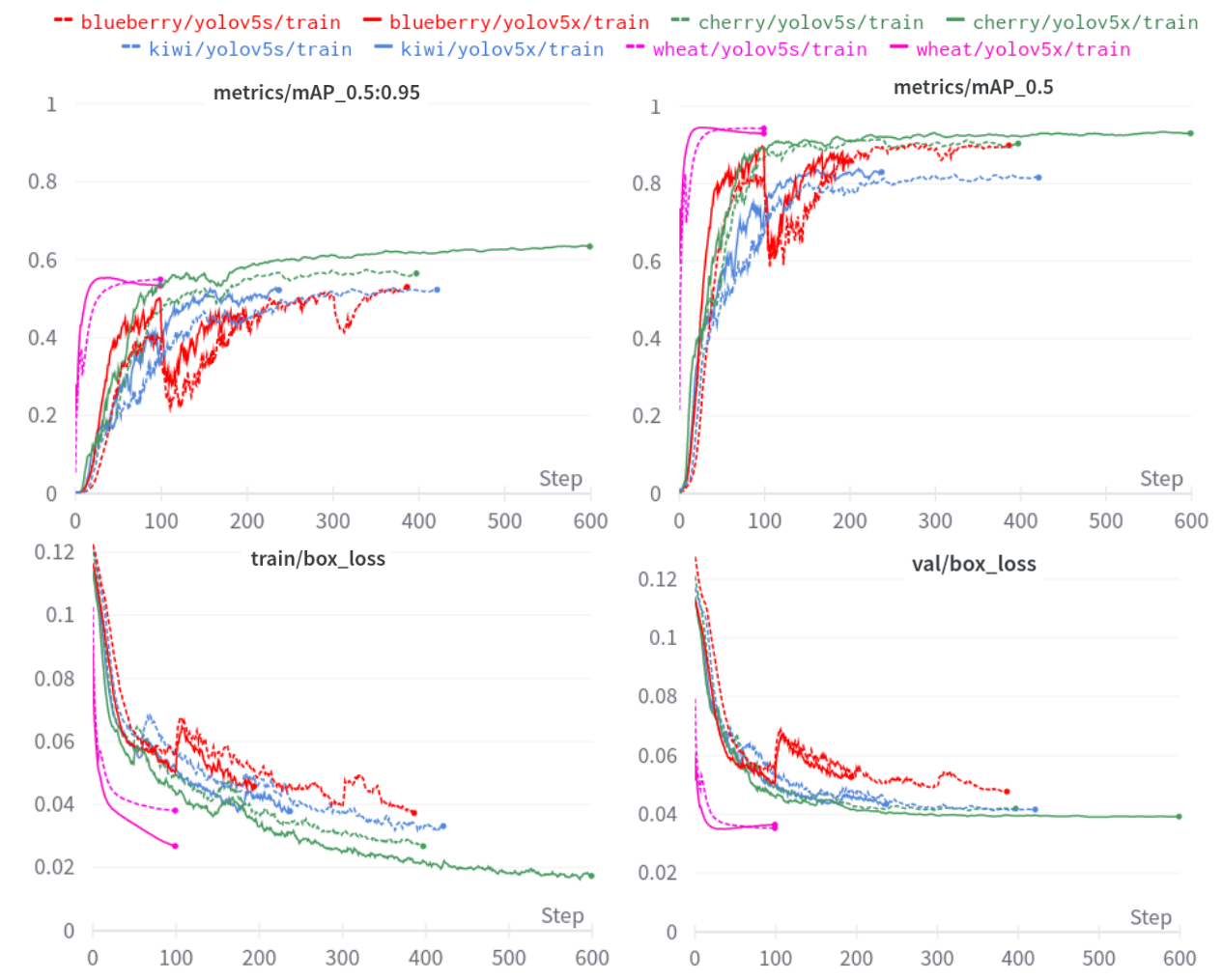}
\caption{mAP performance metrics and train/validation bounding box loss plots for newly added 4 fruits/crops. Due to early-stopping mechanism, each experiment has a varying step length but it should have the same length for its metric and loss.}
\label{fig:loss-plot}
\end{figure}

Train/validation losses are important measures to see the model's performance and behaviours. Figure \ref{fig:loss-plot} reports two mAP metrics and losses results for newly added fruits, blueberry, cherry, kiwi, and wheat. An early-stopping mechanism that terminates the training phase if the model's evaluation has not improved for $N=10$ consecutive epoch/steps was activated, causing different steps for each fruit. All fruits are nicely converged without overfitting and achieve impressive mAP. There are bumps around 100 steps for blueberry (red), and it is a common effect of image augmentation (e.g., probability-based geometry or colour transforms) while training.

\subsection{Qualitative object detection results}
In this section, we demonstrate qualitative 11 fruits detection performance. All images are randomly drawn from test sets (i.e., unseen data while training), and the best performing model, Yolov5x, with RGB images, are utilised for inference. In terms of inference time, it took 4.2$ms$ (238$Hz$) and 10.3$ms$ (97$Hz$) average inference time/image for Yolov5s and Yolov5x models given 640x640 resized image on NVIDIA RTX 3090 GPU respectively. This inference time is matched with what Yolov5 reported \cite{glenn-jocher-2021-5563715} and is sufficient for real-time processing applications.

\begin{figure}
\centering
\includegraphics[width=0.9\textwidth]{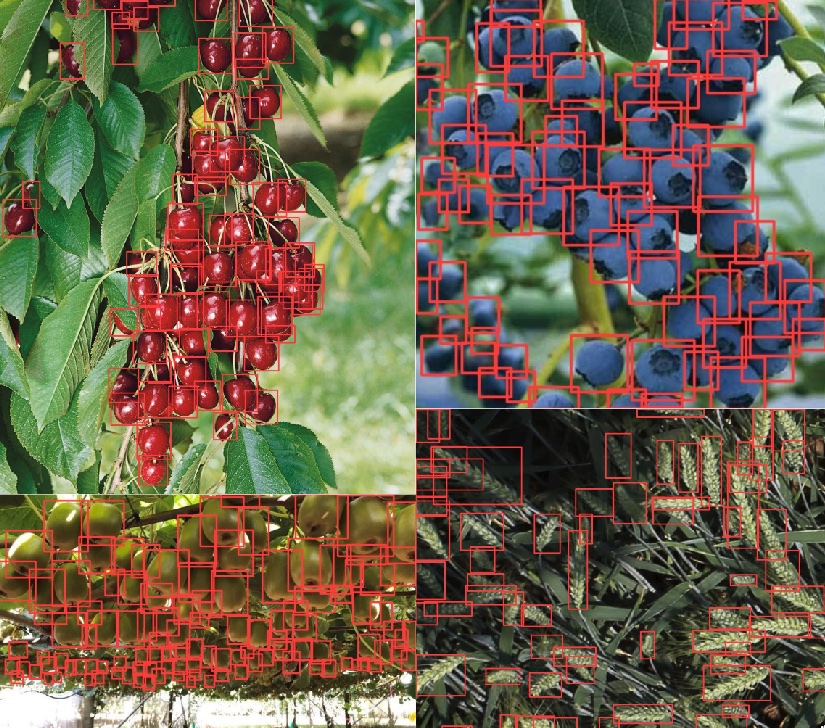}
\caption{Newly added 4 fruits/crop prediction results using Yolov5x and RGB test images. Images are obtained from Google Images and Kaggle wheat detection competition\protect\footnotemark.}
\label{fig:new-fruits-rgb-predic}
\end{figure}

Despite we used the same training dataset of 7 fruits as used in our previous study\cite{Sa2016-zv} in 2015, we can qualitatively observe performance improvement in the state-of-the-art object detector. This can clearly be seen from detecting small scale objects as depicted in Figure \ref{fig:new-fruits-rgb-predic}. Object detection has been actively developed since the early era of deep learning and achieved outstanding performance enhancement in accuracy and inference speed by developing strong image augmentations, model architectures, and hardware and software optimisation. Figure \ref{fig:new-fruits-rgb-predic} exemplify 11 fruits detection results. Due to lengthy pages, we refer to the supplementary material for the all predictions of 11 fruits presented in this paper.

\footnotetext{\url{https://www.kaggle.com/c/global-wheat-detection}}

\begin{figure}[H]
\centering
\includegraphics[width=0.9\textwidth]{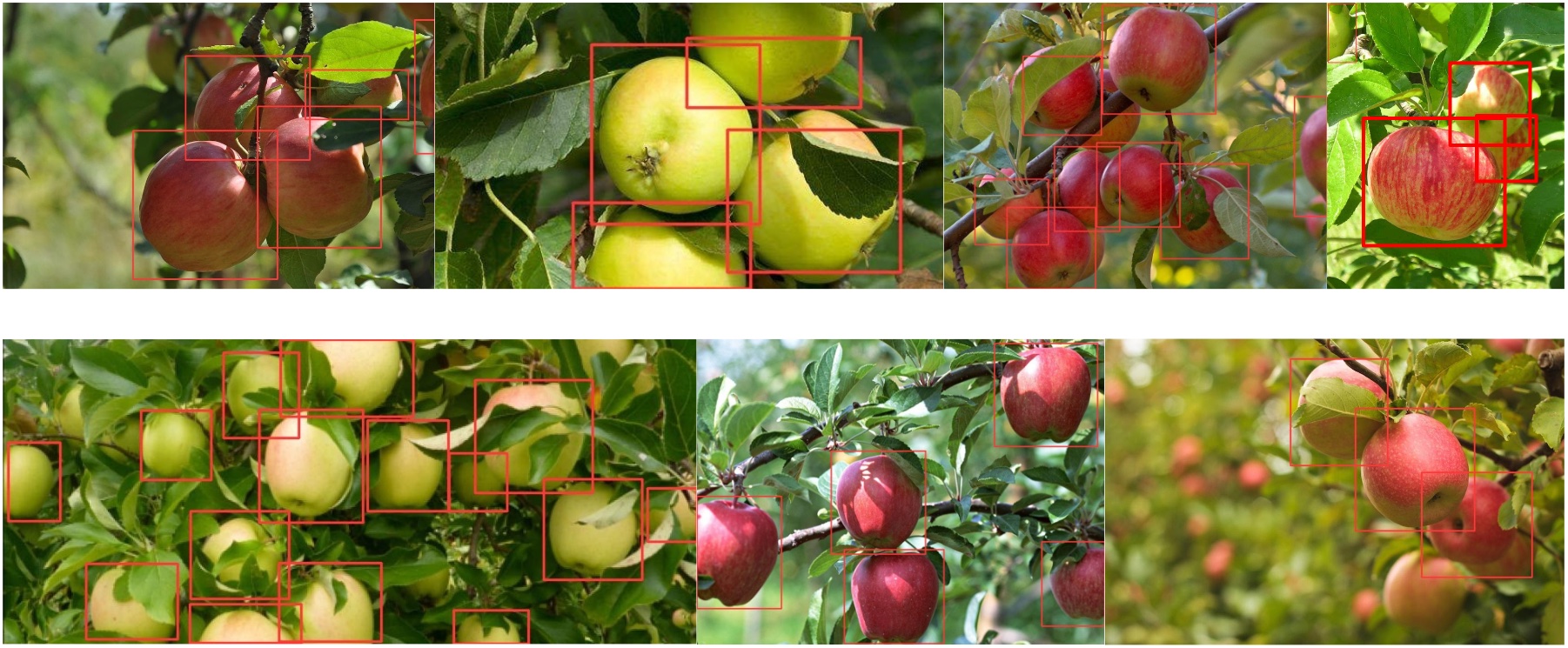}
\caption{Apple prediction results using Yolov5x and RGB test images that achieved 0.78 $\text{mAP}_{\mbox{\tiny{[0.5:0.95]}}}$. Images are obtained from \cite{Sa2016-zv}.}
\label{fig:apple-rgb-predic}
\end{figure}

\begin{figure}[H]
\centering
\includegraphics[width=0.9\textwidth]{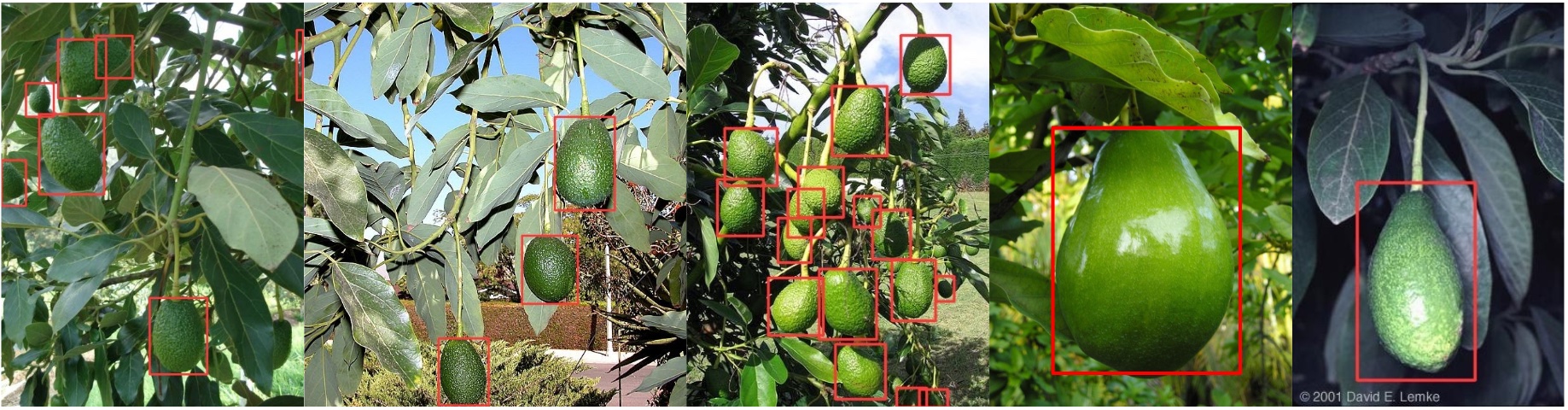}
\caption{Avocado prediction results using Yolov5x and RGB test images that achieved 0.77 $\text{mAP}_{\mbox{\tiny{[0.5:0.95]}}}$. Images are obtained from \cite{Sa2016-zv}.}
\label{fig:avocado-rgb-predic}
\end{figure}

\begin{figure}[H]
\centering
\includegraphics[width=0.9\textwidth]{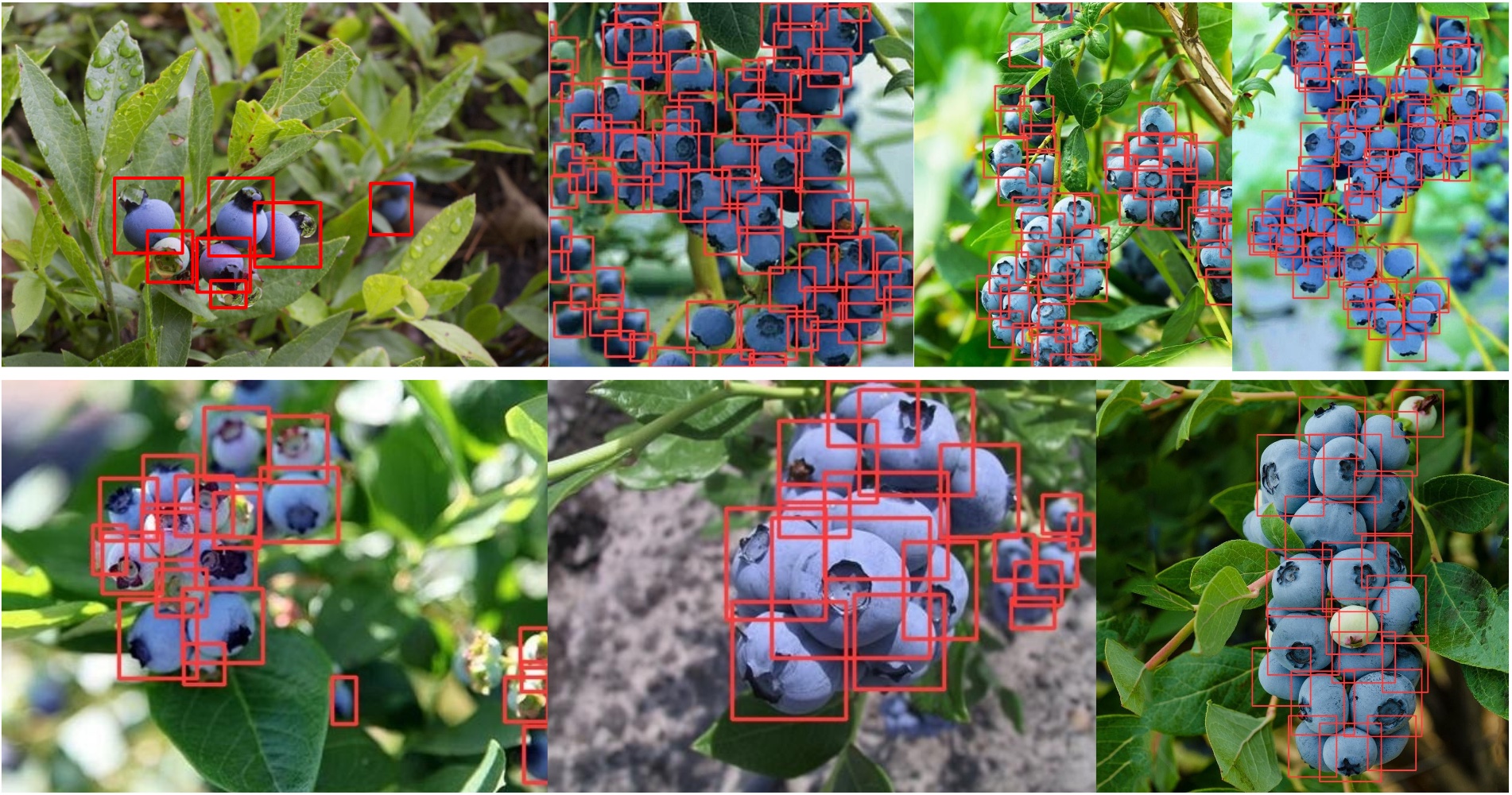}
\caption{Blueberry prediction results using Yolov5x and RGB test images that achieved 0.50 $\text{mAP}_{\mbox{\tiny{[0.5:0.95]}}}$. Blueberry images have relatively high instances/image ratio which yields lower mAP. Images are obtained from Google Images.}
\label{fig:blueberry-rgb-predic}
\end{figure}

\begin{figure}[H]
\centering
\includegraphics[width=0.9\textwidth]{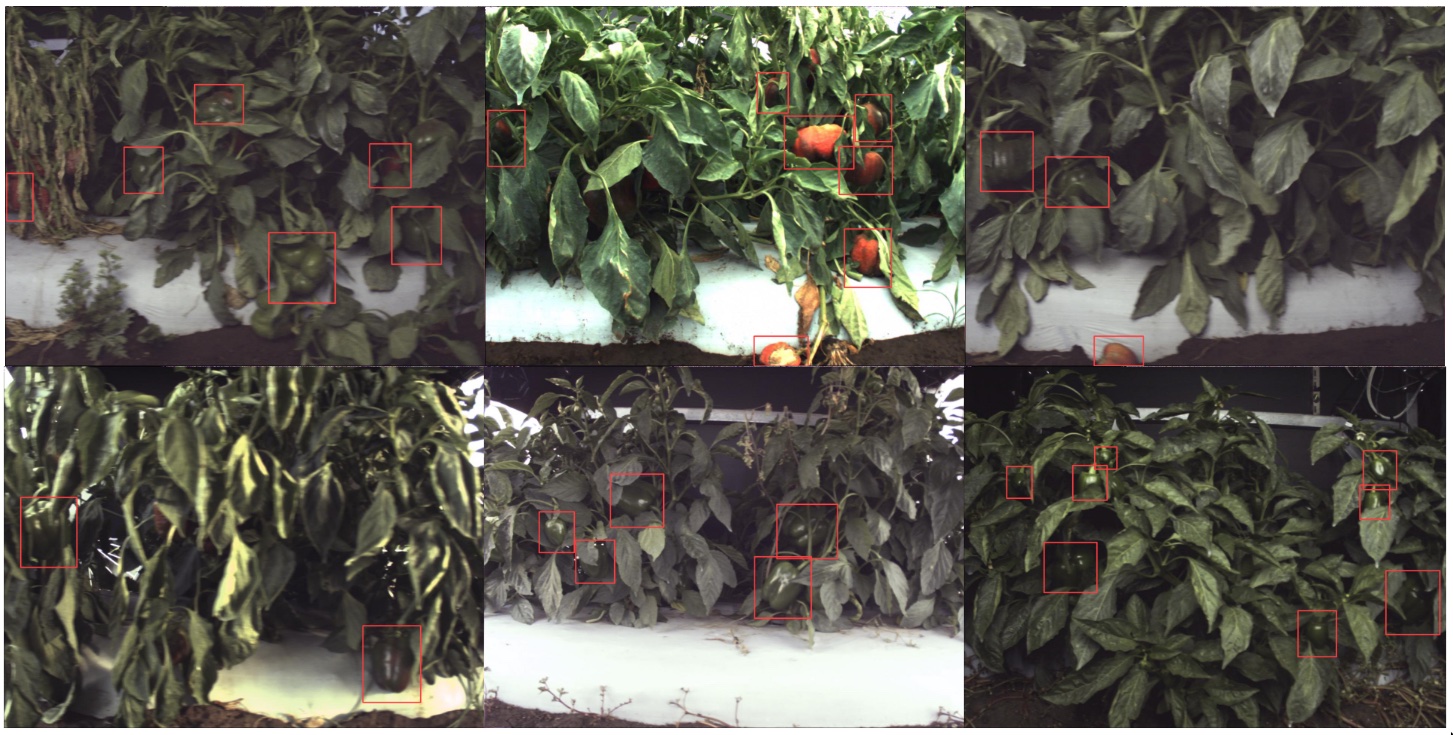}
\caption{Capsicum prediction results using Yolov5x and RGB test images that achieved 0.49 $\text{mAP}_{\mbox{\tiny{[0.5:0.95]}}}$. This is one of the most challenging dataset that collected complex and cluttered real farm environments. Lighting condition is severe, level of occlusion is high, and distance to objects is far. Images are obtained from \cite{Sa2016-zv}.}
\label{fig:capsicum-rgb-predic}
\end{figure}

\begin{figure}[H]
\centering
\includegraphics[width=0.9\textwidth]{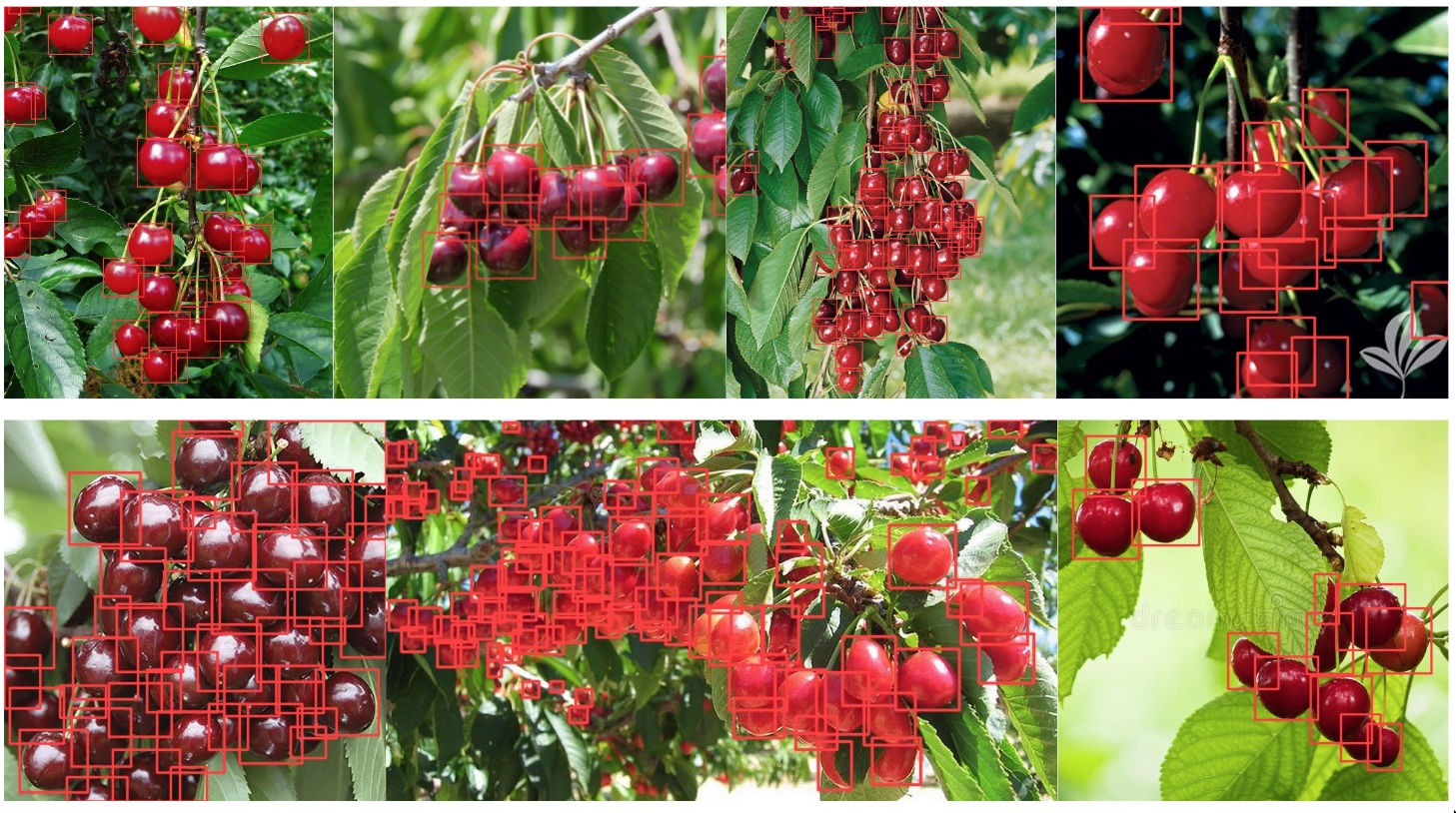}
\caption{Cherry prediction results using Yolov5x and RGB test images that achieved 0.66 $\text{mAP}_{\mbox{\tiny{[0.5:0.95]}}}$. Images are obtained from Google Images.}
\label{fig:cherry-rgb-predic}
\end{figure}

\begin{figure}[H]
\centering
\includegraphics[width=0.9\textwidth]{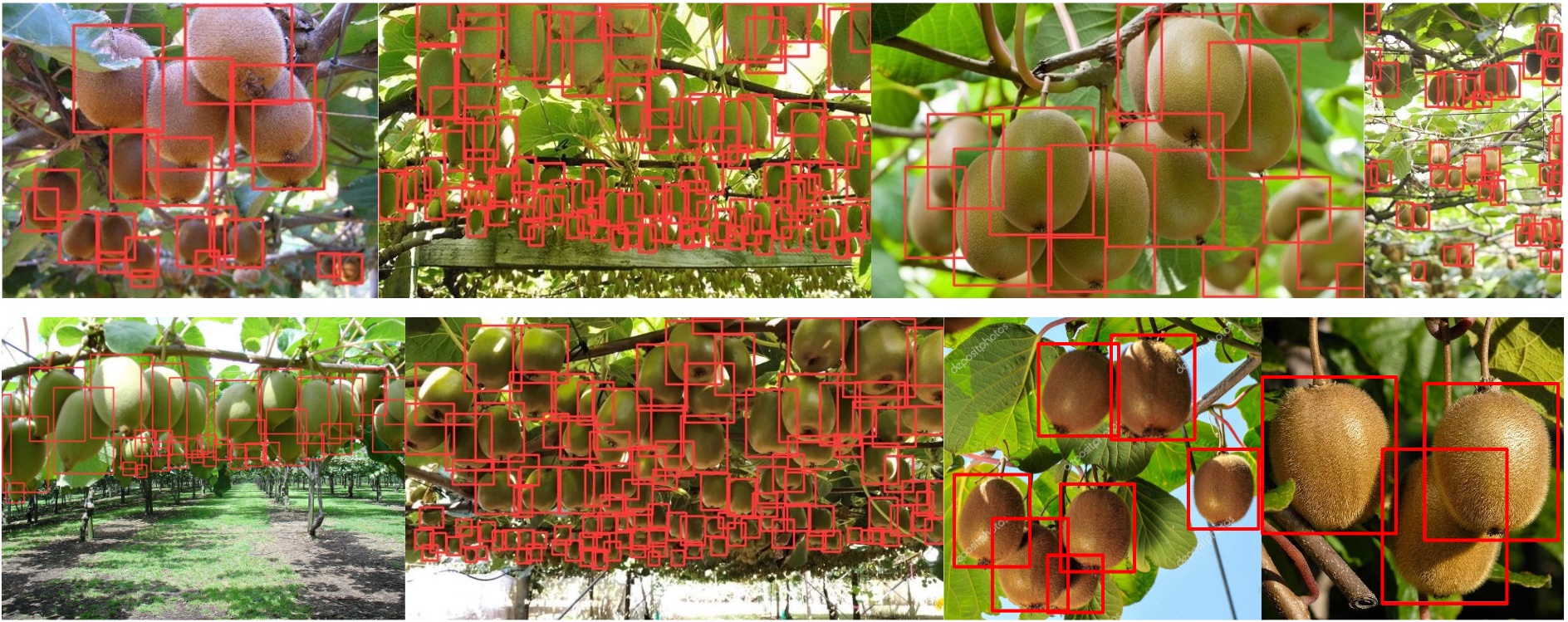}
\caption{Kiwi prediction results using Yolov5x and RGB test images that achieved 0.60 $\text{mAP}_{\mbox{\tiny{[0.5:0.95]}}}$. Images are obtained from Google Images.}
\label{fig:kiwi-rgb-predic}
\end{figure}

\begin{figure}[H]
\centering
\includegraphics[width=0.9\textwidth]{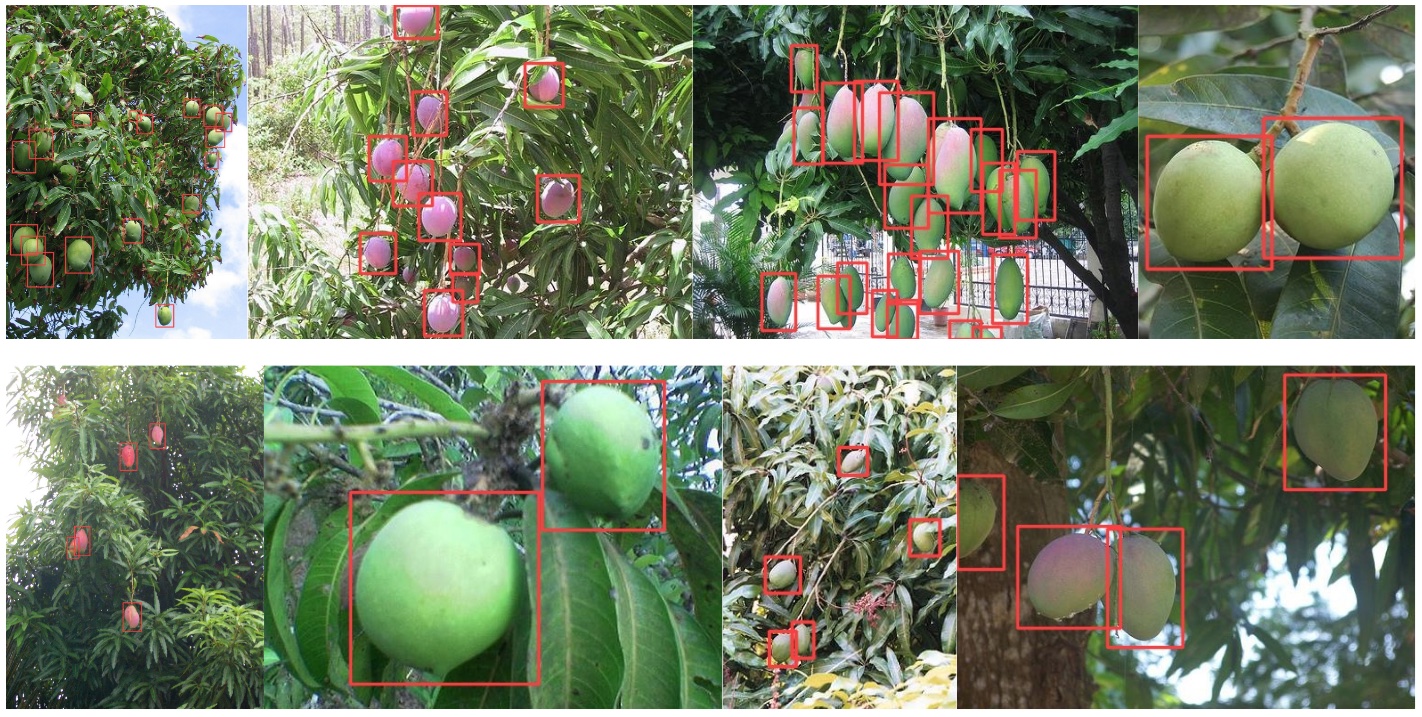}
\caption{Mango prediction results using Yolov5x and RGB test images that achieved 0.69 $\text{mAP}_{\mbox{\tiny{[0.5:0.95]}}}$. Images are obtained from \cite{Sa2016-zv}.}
\label{fig:mango-rgb-predic}
\end{figure}

\begin{figure}[H]
\centering
\includegraphics[width=0.9\textwidth]{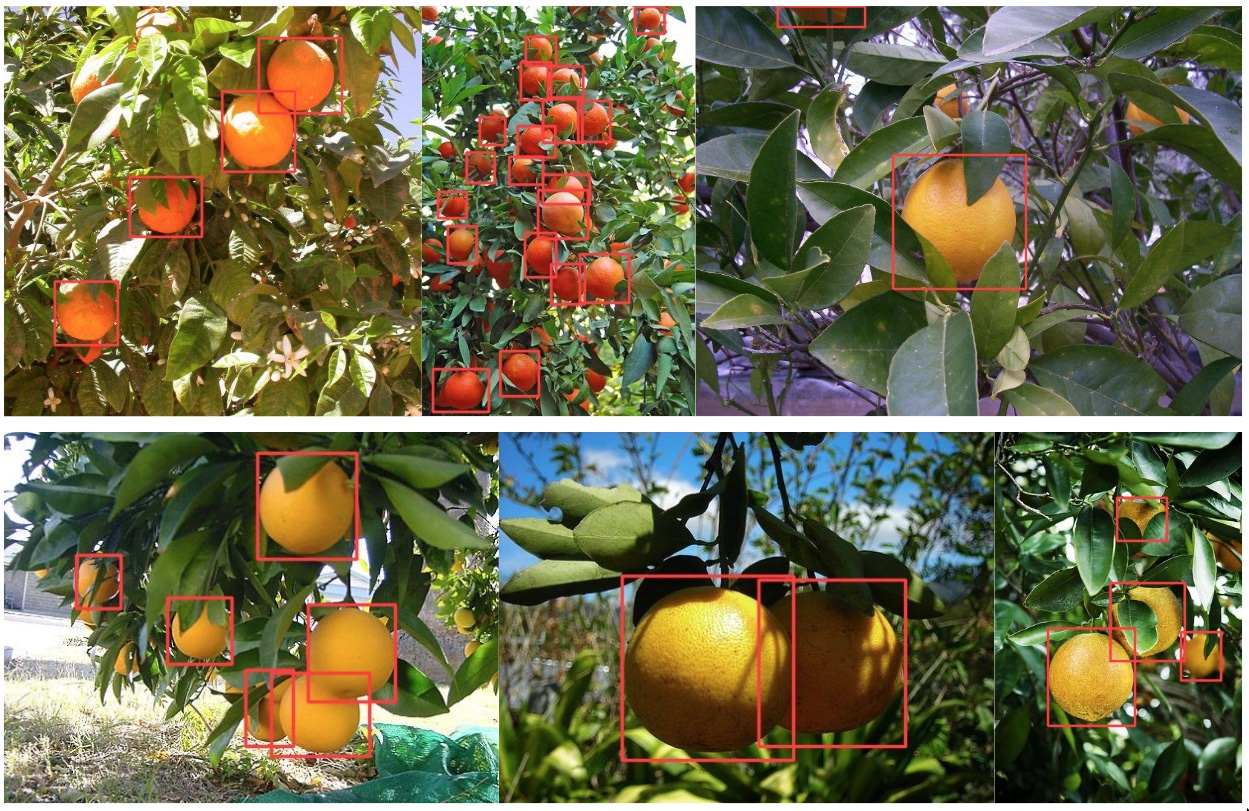}
\caption{Orange prediction results using Yolov5x and RGB test images that achieved 0.73 $\text{mAP}_{\mbox{\tiny{[0.5:0.95]}}}$. Images are obtained from \cite{Sa2016-zv}.}
\label{fig:orange-rgb-predic}
\end{figure}

\begin{figure}[H]
\centering
\includegraphics[width=0.9\textwidth]{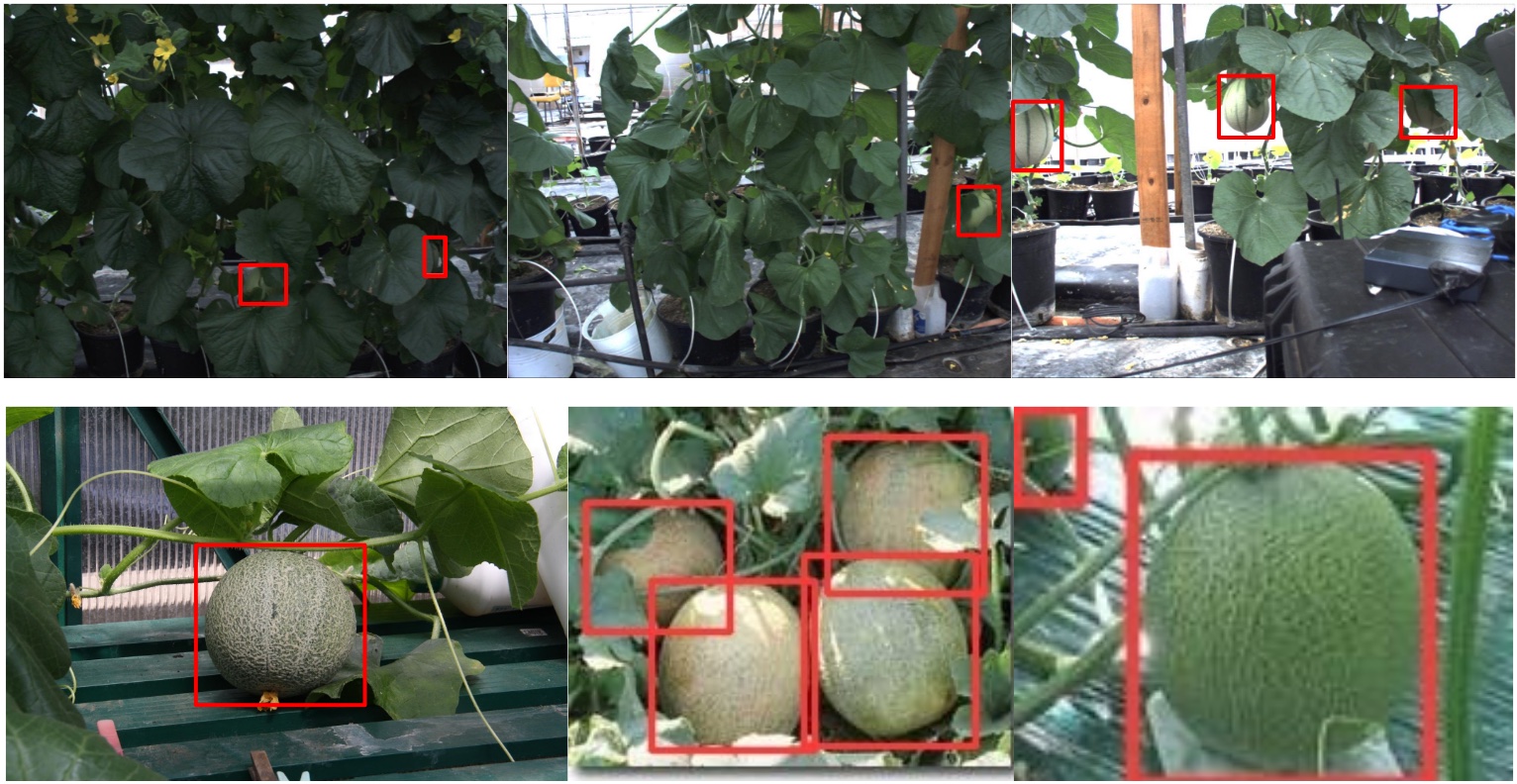}
\caption{Mango prediction results using Yolov5x and RGB test images that achieved 0.69 $\text{mAP}_{\mbox{\tiny{[0.5:0.95]}}}$. Top row images are obtained from \cite{Sa2016-zv} and bottom are from Google Images.}
\label{fig:rockmelon-rgb-predic}
\end{figure}

\begin{figure}[H]
\centering
\includegraphics[width=0.9\textwidth]{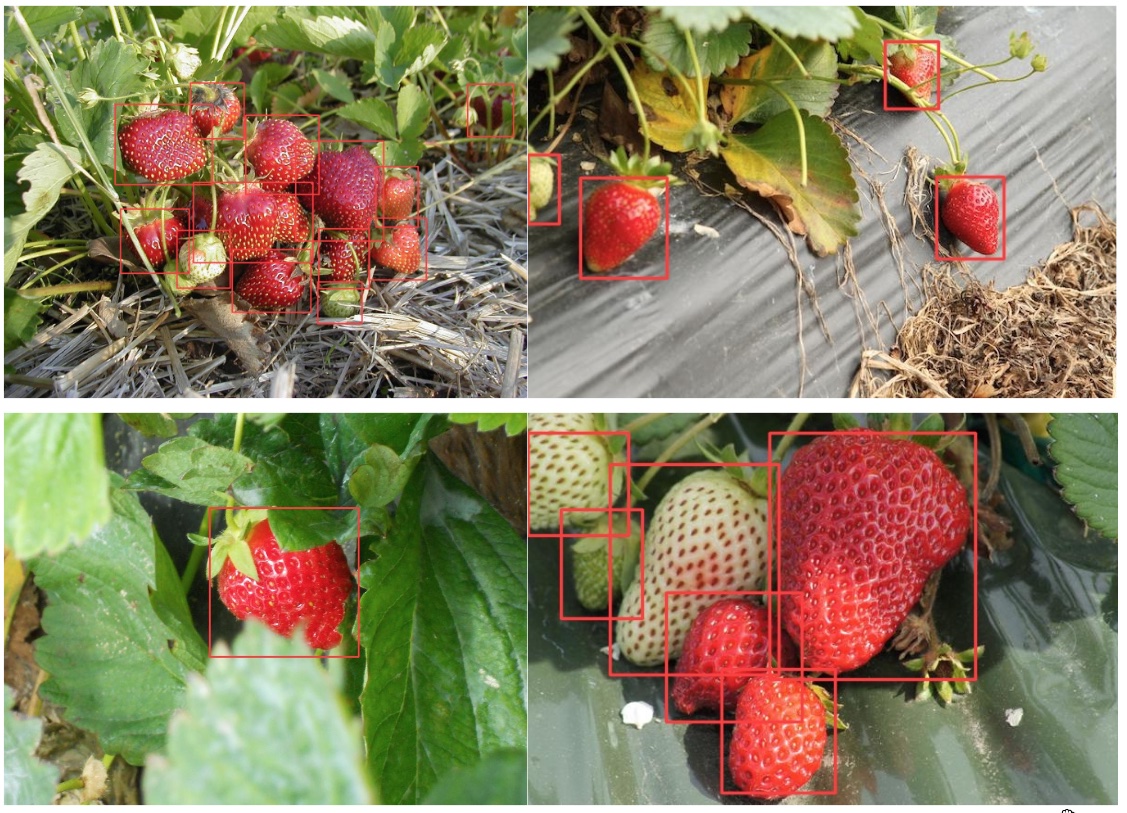}
\caption{Strawberry prediction results using Yolov5x and RGB test images that achieved 0.67 $\text{mAP}_{\mbox{\tiny{[0.5:0.95]}}}$. Images are obtained from \cite{Sa2016-zv}.}
\label{fig:strawberry-rgb-predic}
\end{figure}

\begin{figure}[H]
\centering
\includegraphics[width=0.9\textwidth]{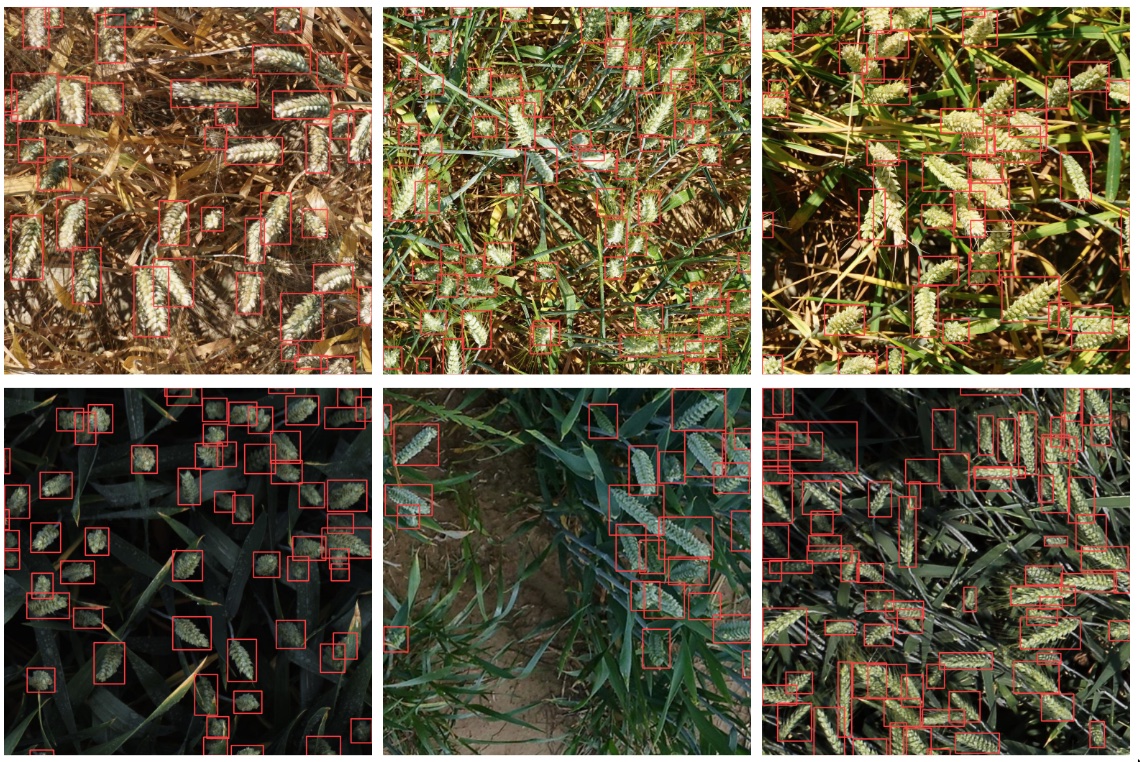}
\caption{Wheat prediction results using Yolov5x and RGB test images that achieved 0.56 $\text{mAP}_{\mbox{\tiny{[0.5:0.95]}}}$ Images are obtained from Kaggle wheat detection competition\protect\footnotemark.}
\label{fig:wheat-rgb-predic}
\end{figure}
\footnotetext{\url{https://www.kaggle.com/c/global-wheat-detection}}

\section{Remaining challenges and limitations}\label{sec:discussion}
\begin{figure}
\centering
\includegraphics[width=\textwidth]{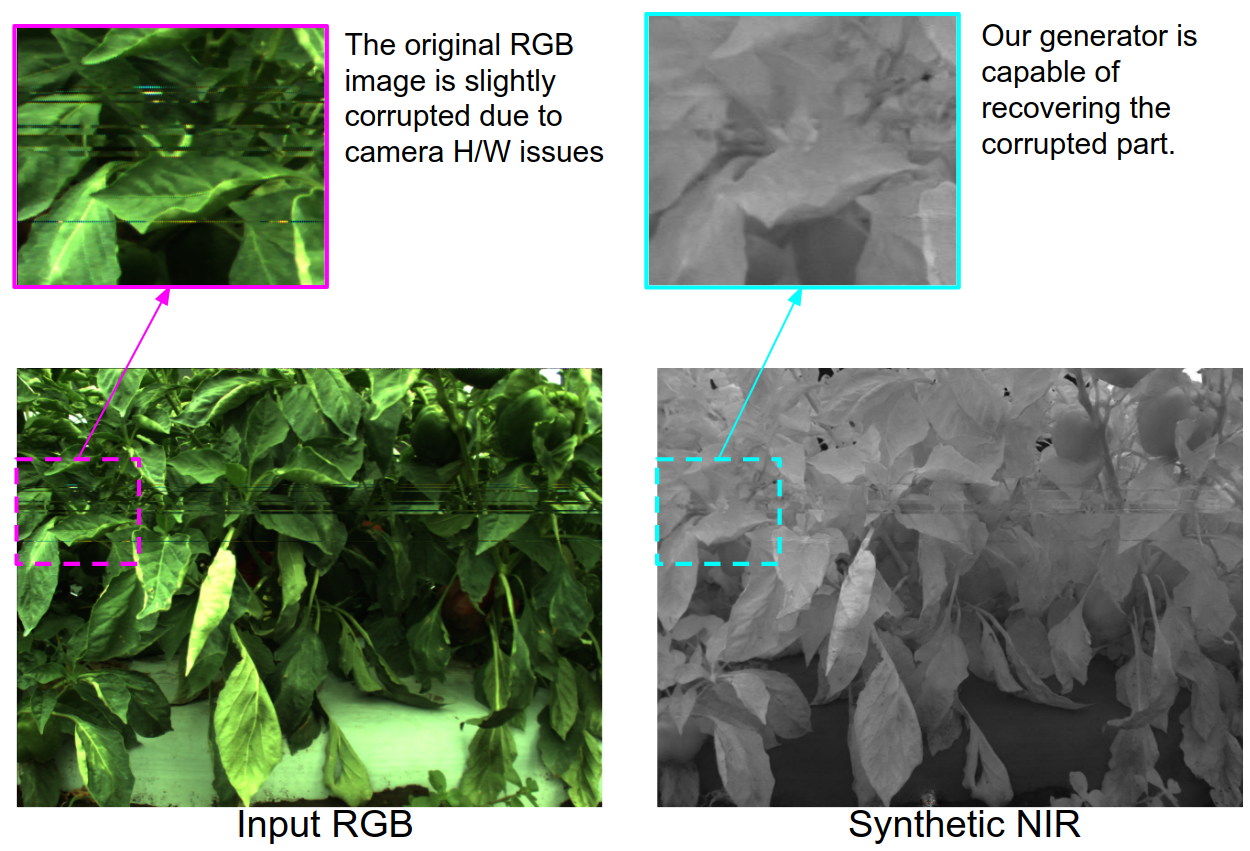}
\caption{This figure illustrates in-painting capabilities of our generator. Left image shows data corruption due to hardware issue and right depicts recovered region of interest (cyan) in synthetic NIR image.}
\label{fig:capsicum-recover}
\end{figure}

While conducting experiments, we observed interesting points and limitations of the proposed approach. Firstly, our synthetic NIR generator can recover small defected data. As shown in Figure \ref{fig:capsicum-recover}, there were a couple of corrupted horizontal lines due to camera hardware issues (e.g., data stream reaches maximum bandwidth of Ethernet interface or abnormally high camera temperature). These artefacts are slightly recovered in the Synthetic NIR because the generator learnt how to incorporate adjacent pixel information to determine NIR pixel value. Eventually, this creates a blurring effect, filling one horizontal line with interpolated data. We agree that it is difficult to argue whether this leads positive or negative impact on the performance. However, if the level of corruption is small (e.g., one or two-pixel rows) and frequently happens, the generator can effectively reject the abnormality.

\begin{figure}
\centering
\includegraphics[width=\textwidth]{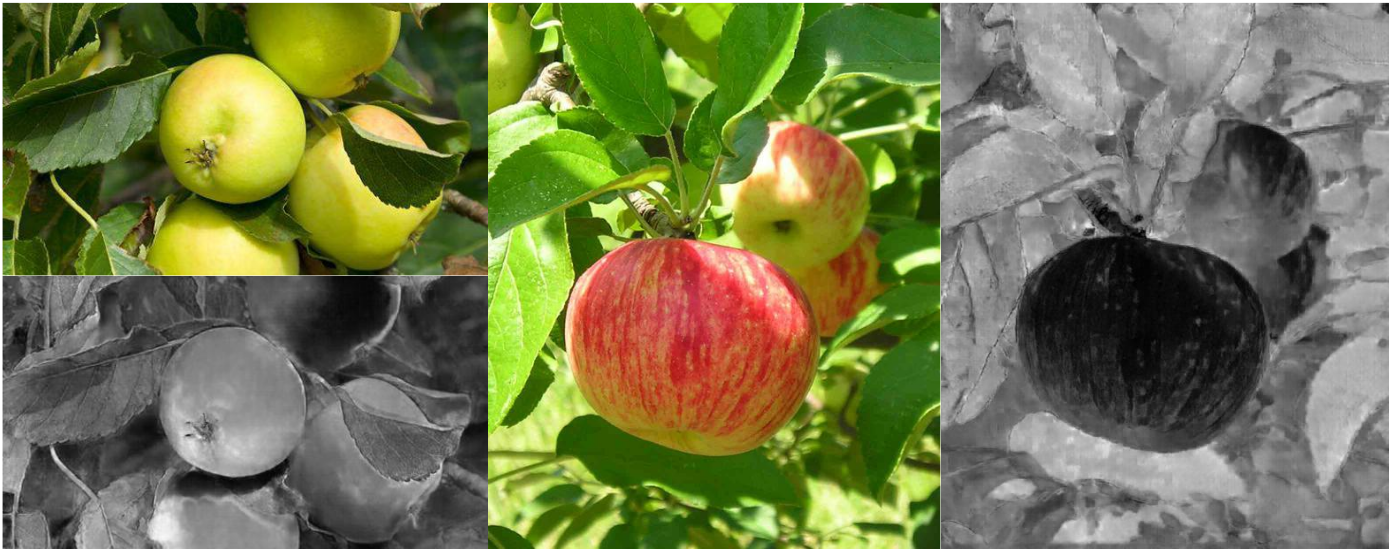}
\includegraphics[width=\textwidth]{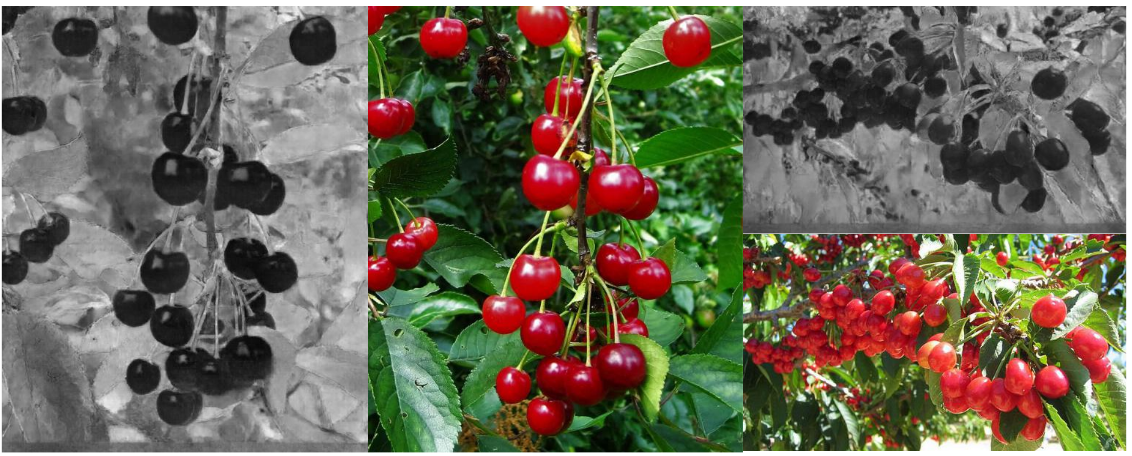}
\includegraphics[width=\textwidth]{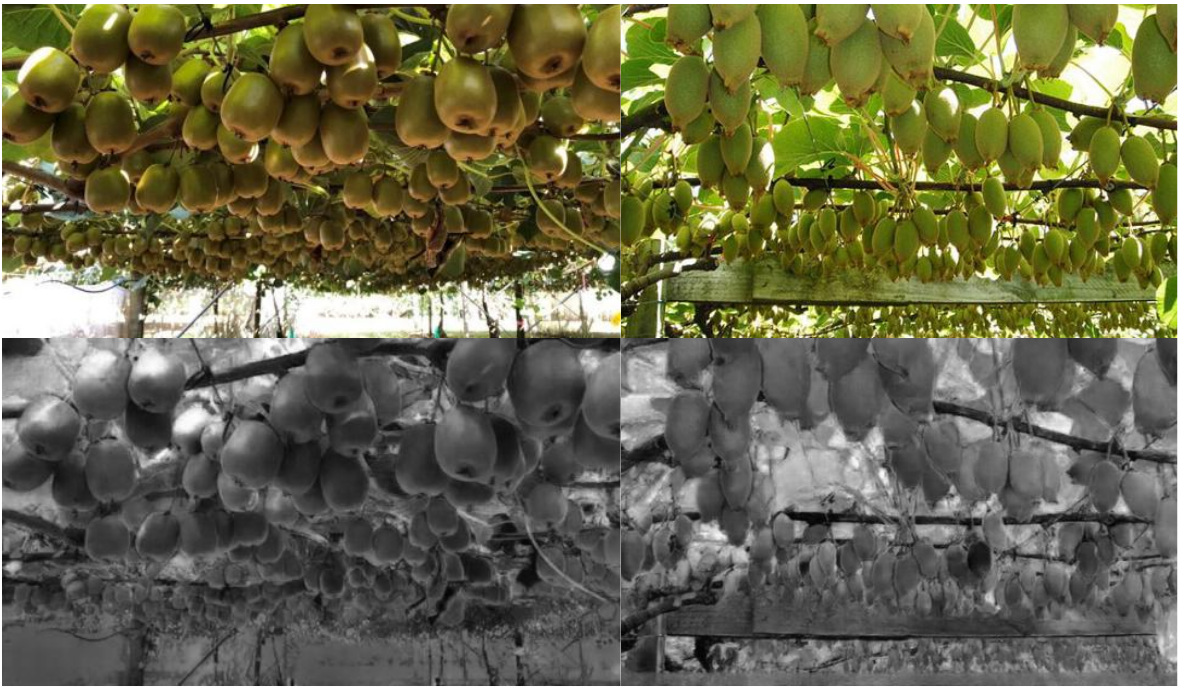}
\caption{Synthetic NIR generation other than capsicum dataset. Each row shows the results for apple, cherry, and kiwi respectively. apple detection performance improved 0.005 $\text{mAP}_{\mbox{\tiny{[0.5:0.95]}}}$ with synthetic NIR. Whereas cherry and kiwi's performance decreased by 0.03 and 0.09 $\text{mAP}_{\mbox{\tiny{[0.5:0.95]}}}$.}
\label{fig:nir-failuer}
\end{figure}

Another discussion point and limitation is a marginal improvement or even degraded performance with synthetic NIR images. Only 4 fruits out of 11 demonstrated superior results with the additional information. The major reason is the large discrepancy between train and test sets. For example, Figure \ref{fig:nir-failuer} shows synthetic NIR images given RGB inputs for apple, cherry, and kiwi. Moreover, these test images were obtained from the Internet to hold high variation properties. Therefore, it is difficult to tell if the generated images are good or poor due to the lack of original NIR images.

\begin{figure}
\centering
\includegraphics[width=\textwidth]{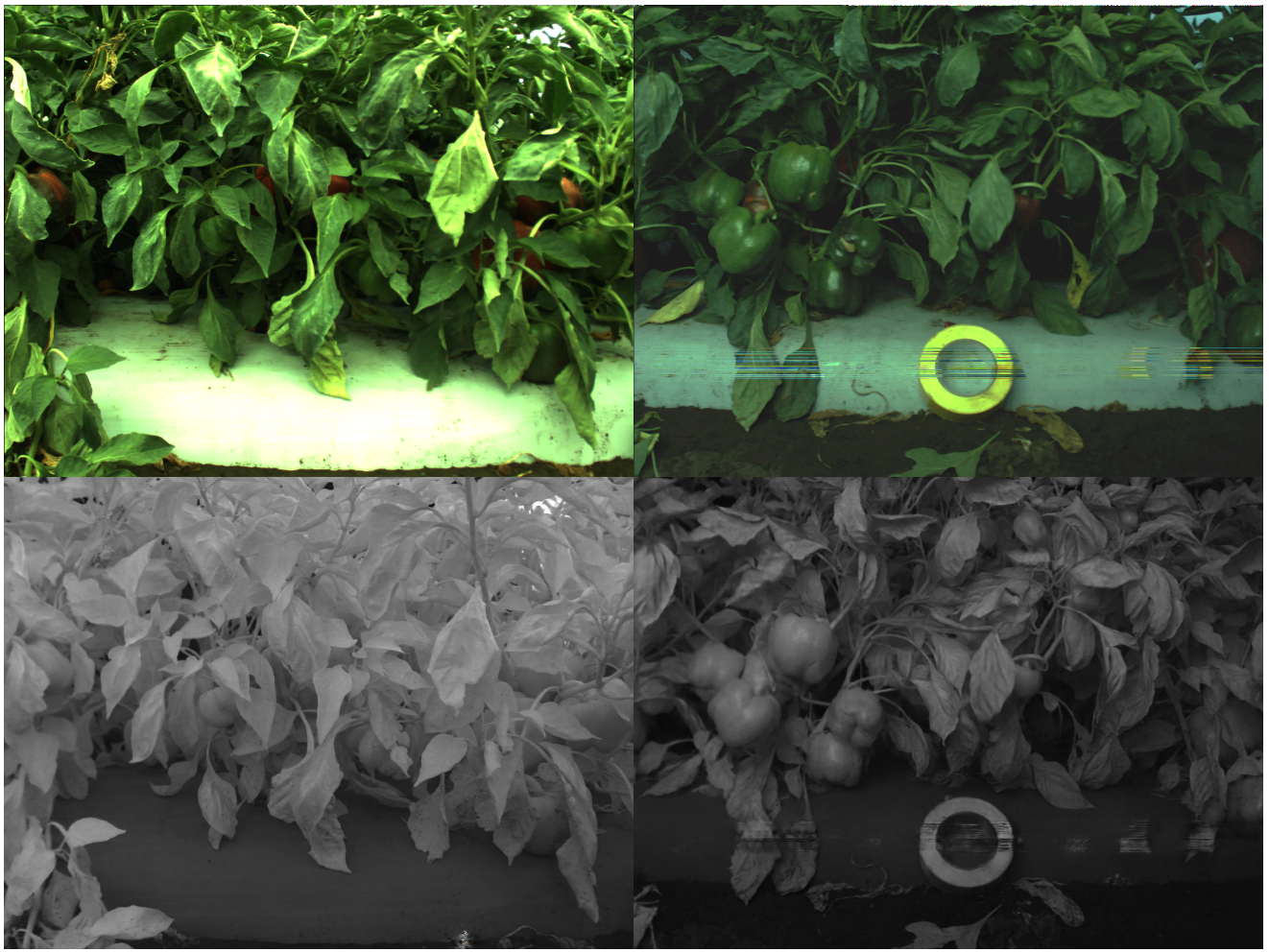}
\caption{Synthetic NIR generation of capsicum dataset. Each row shows RGB and the corresponding NIR. Interestingly, capsicum detection performance decreased 0.01 $\text{mAP}_{\mbox{\tiny{[0.5:0.95]}}}$ with synthetic NIR but increased 0.01 $\text{mAP}_{\mbox{\tiny{[0.5]}}}$. This suggests that synthetic NIR contributed to inaccurately detect instances $\text{IoU} < 0.5$ which can be hidden capsicums under shadow. But these are rejected in evaluating more strict metric, $\text{mAP}_{\mbox{\tiny{[0.5:0.95]}}}$.}
\label{fig:capsicum-success}
\end{figure}

On the other hand, as shown in Figure \ref{fig:capsicum-success}, our generator properly produced synthetic NIR images given test images sampled from a similar distribution of train set. From this experiment, we would like to argue that it is very challenging to generalise our generator model, resulting in faulty and unrealistic samples. However, it should work with samples drawn from similar environments and conditions. Critical properties are consistent lighting and radiometric calibration.

\section{Conclusions and outlook}\label{sec:conclusions}
In this paper, we presented methodologies for generating synthetic NIR images using deep neural networks un-supervised (only required NIR-RGB pair). By adopting 3 public datasets with oversampling, we demonstrated the importance of the scale of the training dataset. It turned out that even with redundant information, it helped to stabilise parameters and led to superior performance. We re-processed these datasets and made them publicly available. These synthetic NIR images are rigorously evaluated with 11 fruits (7 from our previous study and 4 newly added dataset). These are also publicly available in various bounding box formats. This will allow other researchers to use this dataset easily and in a timely way. Early-fusion manner object detection experiments are conducted, and detailed analysis and discussion are shared with readers. 

Although the scale of object annotation is relatively smaller than other giant datasets such as ImageNet or COCO, KITTI, these agriculture and horticulture focused datasets will be useful in many aspects. It can be used for model in-domain-pretraining, a pre-step prior to in-task training (or finetune). For instance, if one wants to train a cherry detector with its own dataset, it makes more sense to pretrain with our dataset rather than ImageNet or COCO, which contain a lot of non-agricultural contexts (e.g., car, buildings, motorcycle, or ship). Another use case is that this small dataset can generate pseudo annotations. Given an unannotated dataset (e.g., 100k kiwi images), one can obtain predictions (i.e., bounding boxes with confidences) using a trained model on a small dataset. Recursive these iterations can improve performance by a large margin over a baseline model\cite{Xie2019-ue}.

To our best knowledge, this paper introduces the most various type of fruit/crops' bounding box annotation dataset at the moment of writing and we hope this is useful for other follow up studies.

\vspace{6pt} 



\authorcontributions{conceptualization, I.S., J.L. and H.A.; methodology, I.S., J.L., and H.A. ; software, I.S.; validation, I.S., and J.L.; resources, I.S. and J.L.; writing--original draft preparation, I.S., J.L. and H.A.; writing--review and editing, I.S., J.L., H.A., B.M.; visualization and data analysis, I.S.}

\funding{This research received no external funding}

\acknowledgments{We would like to thank to nirscene and SEN12MS's public datasets and Kaggle and Roboflow for their assists for hosting fruits bounding box dataset.}

\conflictsofinterest{The authors declare no conflict of interest.}

\newpage

\abbreviations{The following abbreviations are used in this manuscript:\\

\noindent 
\begin{tabular}{@{}ll}
NIR & Near-Infrared\\
GPU & Graphics Processing Unit\\
DN & Digital Number\\
FID & Frechet Inception Distance\\
mAP & mean Average Precision\\
AP & Average Precision\\
ML & Machine Learning\\
NLP & Natural Language Processing\\
NDVI & Normalised Difference Vegetation Index\\
NDWI & Normalised Difference Water Index\\
EVI & Enhanced Vegetation Index\\
LiDAR & Light Detection and Ranging\\
RGB-D & Red, Green, Blue, and Depth\\
RTK-GPS & Real-Time Kinematic Global Positioning System\\
ESA & the European Space Agency\\
GeoTIFF & Geostationary Earth Orbit Tagged Image File Format\\
GAN & Generative Adversarial Networks\\
cGAN & Conditional Generative Adversarial Networks\\
Pix2pix & Pixel to pixel\\
OASIS & You Only Need Adversarial Supervision for Semantic Image Synthesis\\
MRI & Magnetic Resonance Image\\
CTI &  Computed Tomography Image\\
UNIT & Unsupervised image-to-image translation network\\
RPN & Region Proposal Network\\
DNN & Deep-Neural Networks\\
GDS &  Ground Sample Distance\\
YOLOv5 & You Only Look Once\\
BN & Batch-Normalisation\\
ReLU & Rectified Linear Unit\\
CBL & Convolution Batch Normalisation, and Leaky ReLU\\
CSP & Cross Stage Partial\\
SPP & Spatial Pyramid Pooling\\
FPN & Feature Pyramid Networks\\
PAN &  Path Aggregation Network\\
SSD & Single Stage Detectors\\
GIoU & Generalised Intersection Over Union\\
IoU & Intersection Over Union\\
COCO & Common Objects in Context\\
GAM & Generative Adversarial Metric\\
TP & True Positive\\
TN & True Negative\\
FP & False Positive\\
FN & False Negative\\
P & Precision\\
R & Recall\\
MAE & Mean Absolute Error\\
SSIM & Structural Similarity 

\end{tabular}}
\newpage
\reftitle{References}

\externalbibliography{yes}
\bibliography{./bib/myBib.bib}

\end{document}